%% file: latex/00_G_Searcher.tex
\documentclass[11pt]{article}

\usepackage[final]{acl}

\usepackage{times}
\usepackage{latexsym}
\usepackage{soul}
\usepackage[T1]{fontenc}
\usepackage[utf8]{inputenc}

\usepackage{microtype}

\usepackage{inconsolata}

\usepackage{graphicx}

\usepackage{mdframed}
\usepackage{microtype}
\usepackage{graphicx}
\usepackage{fontawesome5}
\usepackage{subcaption}
\usepackage{booktabs} % for professional tables
\usepackage{xcolor}
\usepackage{pifont}
\usepackage{multirow}
\usepackage{array}
\usepackage{hyperref}              % 支持超链接
\usepackage{url}                   % 网址格式化
\usepackage{booktabs}              % 专业质量的表格线
\usepackage{boxedminipage}

\newlength{\kvqaboxsep}
\newlength{\kvqaboxrule}
\usepackage{amsfonts, amssymb, amsmath} % 数学符号和公式
\usepackage{nicefrac}              % 1/2 等紧凑符号
\usepackage{microtype}             % 微排版优化
\usepackage{xcolor}                % 颜色支持
\usepackage{colortbl}              % 表格颜色
\usepackage{multirow}              % 多行单元格
\usepackage[ruled,vlined,linesnumbered]{algorithm2e}
\usepackage{algpseudocode}
\usepackage[export]{adjustbox}     % 调整盒子
\usepackage{threeparttable}        % 三部分表格
\usepackage{enumitem}              % 列表格式
\usepackage{newfloat}              % 新浮动环境
\usepackage{listings}              % 代码列表
\usepackage{caption}               % 标题格式
\usepackage{wrapfig}               % 环绕图形
\usepackage{pifont}                % 图标字体
\usepackage{float}                 % 浮动对象控制
\usepackage{stmaryrd}
\usepackage{bm}                    % 粗体数学符号
\usepackage{placeins}              % 控制浮动对象的位置
\usepackage{graphicx}              % 图形支持
\definecolor{cadetblue}{rgb}{0.372, 0.620, 0.627}
\usepackage{array}
\usepackage{booktabs}
\usepackage{multirow}

\definecolor{mygray}{gray}{.9}
\definecolor{mygreen}{RGB}{93,173,85}
\definecolor{mywarning}{RGB}{233,144,61}

\definecolor{InvBlue}{RGB}{57,84,148}  
\definecolor{FwdGreen}{RGB}{96,129,63}
\definecolor{MutedPurple}{HTML}{703D8A}
\definecolor{MutedRust}{HTML}{8A543D}
\definecolor{DarkBlue}{RGB}{64,101,149}
\definecolor{azure}{rgb}{0.0, 0.5, 1.0}
\definecolor{gray}{rgb}{0.3, 0.3, 0.3}
\definecolor{DarkGreen}{RGB}{42,110,63}
\definecolor{Green}{rgb}{0.0, 0.5, 0.0}       % 绿色
\definecolor{RedOrange}{rgb}{1.0, 0.27, 0.0}  % 红橙色

\newcommand{\blue}[1]{$_{\color{Green}\downarrow #1}$}
\newcommand{\myred}[1]{$_{\color{RedOrange}\uparrow #1}$}

\newcolumntype{x}[1]{>{\centering\arraybackslash}p{#1pt}}
\newcolumntype{I}{!{\vrule width 1pt}} % 定义新的列类型 I，用于加粗竖线

\usepackage[most]{tcolorbox}
\tcbuselibrary{theorems}
\definecolor{lightgray}{gray}{.9}
\definecolor{deepgray}{gray}{.8}
\definecolor{HdrA}{HTML}{D6E2F0} % Tasks (deepest header tint)
\definecolor{HdrB}{HTML}{E3ECF7} % Settings
\definecolor{HdrC}{HTML}{EFF4FB} % Methods (light header)
\definecolor{HdrD}{HTML}{E9EEF5} % GNN/LLM Baselines section bars
\definecolor{HdrE}{HTML}{F4F7FB} % Qwen 3B/7B sub-bars (lightest)
\tcbset{highlight math/.append style={left=0mm,right=0mm,top=0mm,bottom=0mm, colframe=white}}
\definecolor{keywordcolor}{RGB}{178,34,34} % Define FireBrick color for keywords

\definecolor{TagBlue}{HTML}{1F4E79}
\newtcbox{\searchtag}{%
  on line,
  box align=base,
  colframe=TagBlue,
  colback=TagBlue!2!white, 
  coltext=TagBlue,  
  boxrule=0.5pt,
  arc=1.2pt,
  boxsep=0.3pt,
  left=1.0pt,right=1.0pt,top=0.2pt,bottom=0.2pt,
  fontupper=\ttfamily\small
}

\definecolor{TagTeal}{HTML}{1B6B6F}
\definecolor{TagPurple}{HTML}{5B3FA6}
\definecolor{TagRose}{HTML}{8B2E5A}

\newtcbox{\tealtag}{%
  on line,
  box align=base,
  colframe=TagTeal,
  colback=TagTeal!2!white,
  coltext=TagTeal,
  boxrule=0.5pt,
  arc=1.2pt,
  boxsep=0.3pt,
  left=1.0pt,right=1.0pt,top=0.2pt,bottom=0.2pt,
  fontupper=\ttfamily\small
}

\newtcbox{\purpletag}{%
  on line,
  box align=base,
  colframe=TagPurple,
  colback=TagPurple!2!white,
  coltext=TagPurple,
  boxrule=0.5pt,
  arc=1.2pt,
  boxsep=0.3pt,
  left=1.0pt,right=1.0pt,top=0.2pt,bottom=0.2pt,
  fontupper=\ttfamily\small
}

\newtcbox{\rosetag}{%
  on line,
  box align=base,
  colframe=TagRose,
  colback=TagRose!2!white,
  coltext=TagRose,
  boxrule=0.5pt,
  arc=1.2pt,
  boxsep=0.3pt,
  left=1.0pt,right=1.0pt,top=0.2pt,bottom=0.2pt,
  fontupper=\ttfamily\small 
}

\definecolor{TagOrange}{HTML}{D97706}

\newtcbox{\orangetag}{%
  on line,
  breakable,
  tcbox width=auto limited, % allow wrapping within line width
  box align=base,
  colframe=TagOrange,
  colback=TagOrange!2!white,
  coltext=TagOrange,
  boxrule=0.5pt,
  arc=1.2pt,
  boxsep=0.3pt,
  left=1.0pt,right=1.0pt,top=0.2pt,bottom=0.2pt,
  fontupper=\ttfamily\small
}

\newtcolorbox{casebox}[1]{
  enhanced,
  colback=black!2,
  colframe=black!60,
  boxrule=0.6pt,
  arc=2mm,
  outer arc=2mm,
  top=10mm,
  bottom=4mm,
  left=8mm,
  right=8mm,
  fontupper=\ttfamily\small,
  width=\textwidth,
  title={#1},
  overlay unbroken={
    \path[fill=black!75, draw=black!75]
      (frame.north west) rectangle ([yshift=-18pt]frame.north east);
    \node[anchor=west, font=\bfseries\ttfamily\large, text=white]
      at ([xshift=10pt, yshift=-9pt]frame.north west)
      {\tcbtitle};
  }
}

\newtcolorbox{caseboxpurple}[1]{%
  enhanced,
  colback=violet!5,          % 背景：很浅的紫
  colframe=violet!50!black,  % 边框：稍深一点的紫
  boxrule=0.6pt,
  arc=2mm,
  outer arc=2mm,
  top=12mm,
  bottom=4mm,
  left=8mm,
  right=8mm,
  fontupper=\ttfamily\small, % 里面还是等宽字体
  width=\textwidth,
  title={#1},
  overlay unbroken={
    \path[fill=violet!70!black, draw=violet!70!black]
      (frame.north west) rectangle ([yshift=-18pt]frame.north east);
    \node[anchor=west, font=\bfseries\ttfamily\large, text=white]
      at ([xshift=10pt, yshift=-9pt]frame.north west)
      {\tcbtitle};
  }
}

\makeatletter
\newcommand{\thickhline}{%
    \noalign {\ifnum 0=`}\fi \hrule height 1pt
    \futurelet \reserved@a \@xhline
}
\makeatother



\newcommand{\cmark}{\ding{51}}%
\usepackage[capitalize]{cleveref}
\crefname{proposition}{Prop.}{Props.}
\crefname{section}{Sec.}{Secs.}
\crefname{table}{Tab.}{Tabs.}

\usepackage{xspace}
\makeatletter
\DeclareRobustCommand\onedot{\futurelet\@let@token\@onedot}
\def\@onedot{\ifx\@let@token.\else.\null\fi\xspace}

\definecolor{CadetBlue}{RGB}{95,158,160} % Define CadetBlue color
\makeatother

\usepackage{amsmath}

\usepackage{amssymb}
\usepackage{mathtools}
\usepackage{amsthm}

\theoremstyle{plain}
\newtheorem{theorem}{Theorem}[section]

\theoremstyle{definition}
\newtheorem{definition}[theorem]{Definition}

\theoremstyle{remark}
\newtheorem{remark}[theorem]{Remark}
\crefname{definition}{Definition}{Definitions}
\Crefname{definition}{Definition}{Definitions}

\usepackage[textsize=tiny]{todonotes}

\title{AgentGL: Towards Agentic Graph Learning with LLMs via \\Reinforcement Learning}

\author{
  Yuanfu Sun$^{1,2}$\footnotemark[1], Kang Li$^{3}$\footnotemark[1], Dongzhe Fan$^{1,2}$, Jiajin Liu$^{1,2}$, Qiaoyu Tan$^{1}$\footnotemark[2] \\
  $^1$New York University Shanghai 
  $^2$New York University
  $^3$Tsinghua University \\ 
  \texttt{\{yuanfu.sun, qiaoyu.tan\}@nyu.edu, lik24@mails.tsinghua.edu.cn}\\
}
\begin{document}
\maketitle
\renewcommand{\thefootnote}{\fnsymbol{footnote}}
\footnotetext[1]{Equal contribution} \footnotetext[2]{Corresponding author}

% 如果 01_abstract.tex 里已经包含 \begin{abstract}...\end{abstract}，
% 这里直接 \input 即可；否则你可以改成 \begin{abstract}\input{01_abstract}\end{abstract}
\input{latex/01_abstract}

\input{latex/02_introduction}

\input{latex/03_relatedwork}
\input{latex/04_problem_state}

\input{latex/05_methodology}

\input{latex/06_experiments}

\input{latex/07_conclusion}

% 如果你在 07_appendix.tex 里不写 \appendix，可以在这里统一切换附录模式
\bibliography{custom}

\appendix
\input{latex/08_appendix}

\end{document}

%% file: latex/01_abstract.tex
\begin{abstract}
Large Language Models (LLMs) increasingly rely on agentic capabilities—iterative retrieval, tool use, and decision-making—to overcome the limits of static, parametric knowledge. Yet existing agentic frameworks treat external information as unstructured text and fail to leverage the topological dependencies inherent in real-world data. To bridge this gap, we introduce Agentic Graph Learning (AGL), a paradigm that reframes graph learning as an interleaved process of topology-aware navigation and LLM-based inference. Specifically, we propose AgentGL, the first reinforcement learning (RL)–driven framework for AGL. AgentGL equips an LLM agent with graph-native tools for multi-scale exploration, regulates tool usage via search-constrained thinking to balance accuracy and efficiency, and employs a graph-conditioned curriculum RL strategy to stabilize long-horizon policy learning without step-wise supervision. Across diverse Text-Attributed Graph (TAG) benchmarks and multiple LLM backbones, AgentGL substantially outperforms strong GraphLLMs and GraphRAG baselines, achieving absolute improvements of up to 17.5\% in node classification and 28.4\% in link prediction. These results demonstrate that AGL is a promising frontier for enabling LLMs to autonomously navigate and reason over complex relational environments. The code is publicly available at \url{ https://github.com/sunyuanfu/AgentGL}.
\end{abstract}

%% file: latex/02_introduction.tex
\section{Introduction}
Large Language Models (LLMs) have achieved strong performance across NLP tasks through their broad linguistic and reasoning capabilities \cite{achiam2023gpt,yang2025qwen3}. Yet their parametric knowledge alone is insufficient for many specialized or fast-evolving domains \cite{lewis2020retrieval}. To bridge this gap, Retrieval-Augmented Generation (RAG) \cite{gao2023retrieval} and more recent agentic search frameworks \cite{li2025search,jin2025search,chen2025learning} allow LLMs to iteratively query external resources and integrate retrieved evidence into a dynamic chain of thought.

Despite the power of agentic paradigms, they mainly operate on unstructured text, overlooking the relational structures that underpin many corpora. In critical domains such as citation networks \cite{yang2016revisiting}, social platforms \cite{hamilton2017inductive}, and commercial ecosystems \cite{shchur2018pitfalls}, information naturally manifests as Text-Attributed Graphs (TAGs), where meaning is derived from the interplay between textual content and graph topology. Consequently, agentic systems that rely solely on lexical similarity cannot harness these structural dependencies. This raises a central question: \textit{Can the agentic learning paradigm be extended to graph-structured environments to enable dynamic, topology-aware reasoning, and how can such a system be built efficiently?}

Existing graph learning efforts only partially address this need. Traditional GNNs \cite{kipf2016semi,velickovic2017graph} model structural signals but struggle with rich textual semantics \cite{yan2023comprehensive}. Recent LLM-based Graph Models (GraphLLMs) integrate LLMs with graph information via graph-guided prompting or instruction tuning (e.g., GraphGPT \cite{tang2023graphgpt}, GraphICL \cite{sun2025graphicl}), but these models rely on static graph context extracted once at inference time, preventing adaptive exploration. GraphRAG systems \cite{jimenez2024hipporag,dong2025youtu} construct large text-enriched knowledge graphs (KGs) from corpora, yet these reconstructed KGs are costly to build and do not preserve the native topological correlations present in real TAGs. Consequently, neither GraphLLMs nor GraphRAG offers mechanisms for dynamic evidence acquisition over real-world graph structure.

This motivates the emergence of Agentic Graph Learning (AGL), a new direction where a LLM agent can autonomously navigate a graph, accumulate structural evidence, and iteratively refine its search trajectory based on on-the-fly reasoning.
However, realizing AGL is non-trivial due to two fundamental challenges. \textit{\textbf{(C1) Topology-aware navigation.}} Evidence on a graph is multi-scale: some clues appear in tightly local neighborhoods, whereas others emerge only through broader structural patterns. An agent must decide where to go next in a combinatorial space while avoiding redundant or uninformative regions. \textit{\textbf{(C2) Long-horizon policy optimization.}} Effective graph reasoning frequently requires multi-step exploration, but ground-truth search trajectories are rarely available. This makes it difficult to learn policies that balance exploration, exploitation, and reasoning depth, and easy for agents to drift into irrelevant branches or incur unnecessary tool calls. Addressing these challenges demands a principled formulation of graph-native action spaces and stable training mechanisms for long-horizon decision-making.

% \textbf{\textit{C1: The disorientation dilemma amidst complex graph structures.}} 
% Web-search agents jumps enabled by semantic matching, regardless of the distance and relationship between information sources.
% In contrast, graph-native search is topology-bound and inherently multi-granular: evidence is stratified, residing either in local neighborhoods or global structural patterns.
% Navigating this heterogeneous environment without innate mechanisms to disentangle local interactions from global structural priors, the LLM struggles to determine where to initiate the search and how to traverse the topology, often becoming disoriented in the combinatorial explosion of connections.
% % Challenge 2: The Policy/Strategy Problem
% \textbf{\textit{C2: The deficit in strategic reasoning and policy optimization.}}
% Effective GL is not merely about input-time retrieval, but requires a coherent, interleaved thinking and searching process.
% Current paradigms lack the mechanisms to optimize such long-horizon trajectories.
% Without a learned policy to regulate its behavior, the LLM struggles to balance competing objectives: it must deepen reasoning to ensure accuracy while simultaneously constraining its cognitive budget to maintain efficiency.
% Optimizing this trade-off is particularly challenging due to the scarcity of step-by-step supervision (ground-truth paths), leaving agents prone to error cascading and suboptimal planning.

% \qiaoyu{However, xxx is challenging to develop an effective Search-Augmented LLM for graph learning due to two hurdles. First}
To address these challenges, we propose \texttt{AgentGL}, a framework that formulates graph learning as an agentic decision-making process optimized through reinforcement learning (RL). \texttt{AgentGL} equips LLM  with a suite of graph-native search tools, including local neighborhood expansion, hop-constrained traversal, and global evidence probing that enable multi-scale structural exploration tailored to the task. To prevent over-searching and encourage deeper reasoning on retrieved evidence, we introduce search-constrained thinking, a mechanism that biases the LLM agent toward reflective inference before invoking additional graph queries. To support stable long-horizon learning without step-by-step trajectory supervision, we further develop a graph-conditioned curriculum RL strategy that progressively increases topology exploration difficulty, integrates multi-faceted rewards, and enforces efficient use of graph tools under limited budgets. Together, \texttt{AgentGL} enables LLM agents to learn adaptive, topology-aware search policies that significantly enhance performance on diverse graph reasoning tasks.

% In light of these imperatives, we present \textit{\textbf{\texttt{AgentGL}}}, the first framework that reframes GL as an agentic decision-making process driven by Reinforcement Learning (RL). 
% \texttt{AgentGL} empowers the LLM to function as a strategic graph navigator, actively traversing the topology to construct its own evidence path.
% To address \textit{\textbf{C1}}, we equip the LLM with a suite of Graph-Native Search Tools, enabling the agent to seamlessly span local context and global structure, ensuring evidence is captured across all topological scales.
% Beyond mere perception, to tackle \textit{\textbf{C2}}, we enforce a "Think more, Search less" cognitive paradigm via \textit{Search-Constrained Thinking}; this mechanism compels the LLM to perform deep reasoning on searched evidence before initiating further exploration, transforming inference from a shallow walk into a rigorous investigation.
% Finally, to eliminate the reliance on expensive step-by-step supervision, we optimize the agent's policy using RL stabilized by Graph-Conditioned Curriculum Learning, enabling the system to autonomously learn to balance exploration efficiency with reasoning accuracy across complex environments.The primary contributions can be summarized as follows.

\begin{itemize}[leftmargin=*, topsep=0mm,label=\ding{70}]
  \item We study Agentic Graph Learning (AGL), a new paradigm that treats graph learning as an interleaved process of topology-aware exploration and LLM-based reasoning. This formulation unifies graph structure, text semantics, and agentic decision-making under a single framework.
  %\textbf{Problem Formulation.} We formally define Agentic Graph Learning, pioneering a shift from static inference to a interleaved process of graph-native searching and reasoning. This formulation unifies structural perception with dynamic evidence acquisition, bridging the chasm between GL and agentic search.
  \item We propose \texttt{AgentGL}, the first RL-driven AGL framework that synergizes \emph{structural perception}, \emph{strategic reasoning}, and \emph{policy learning}. Specifically, it orchestrates graph-native search tools and search-constrained thinking to navigate complex topologies, employing graph-conditioned curriculum-based RL to optimize the policy without step-wise supervision.

  \item We evaluate \texttt{AgentGL} across multiple TAG benchmarks and graph tasks, demonstrating strong improvements over leading GraphLLM and GraphRAG baselines. Specifically, it delivers absolute accuracy improvements of up to \textbf{17.5\%} in node classification and up to \textbf{28.4\%} in link prediction across diverse LLM backbones.
\end{itemize}

% \qiaoyu{To bridge the gap, we propose G-Searcher.}

% \qiaoyu{It is worth noting that our study is fundamental different from well established GraphRAG works. GraphRAG xxxx. One close work to us is GraphCoT, however, it is mainly designed for knowledge graph-based common sense QA tasks, which xxx. Moreover, G-Searcher xxxx. Our key contributions are summarized below. }

%% file: latex/03_relatedwork.tex
\section{Related Work}

\noindent\textbf{Graph Learning with LLMs.}
Recent work has focused on bridging the gap between graph-structured data and LLMs to facilitate graph reasoning.
One line textualizes local structures into natural-language descriptions to support LLM reasoning and contextualized representations~\citep{zhao2023graphtext,guo2023gpt4graph,chen2024exploring,li2024similarity,shi2024retrieval,fang2024gaugllm,he2023harnessing}.
Another line derives graph tokens or structure-aware embeddings and injects them into prompts for graph instruction tuning~\citep{tang2023graphgpt,chenllaga,liu2024can, sun2026mario, zhang2025trustglm}, or performs training-free inference via graph in-context learning ~\citep{sun2025graphicl,huang2023can, liu2024moleculargpt}.
Despite progress, these pipelines are largely static, limiting adaptation when additional evidence is needed during inference time.

\noindent\textbf{Grounding LLMs with External Knowledge.}
While standard RAG improves factuality via static retrieval~\citep{lewis2020retrieval,gao2023retrieval}, agentic search advances this by enabling iterative reasoning through reinforcement learning~\citep{jin2025search,song2025r1,chen2025learning} or prompting~\citep{yao2022react,press2023measuring,li2025search}. However, these methods predominantly target unstructured text. To incorporate structure, GraphRAG approaches~\citep{he2024g,jimenez2024hipporag,dong2025youtu,han2024retrieval} retrieve evidence from graph-structured data. Yet, they often rely on synthetic graphs reconstructed from flat corpora and their task objectives are fundamentally different from GL (\ref{app:graphrag-vs-agl}). Even native-graph methods remain limited: GraphCoT~\citep{jin2024graph} focuses on graph QA, while GraphSearch~\citep{liu2026graphsearch} targets graph learning, yet both rely on heuristic prompting with limited optimization, often yielding sub-optimal solutions.

%% file: latex/04_problem_state.tex
\section{Problem Statement}

\begin{figure*}
    \centering
    \includegraphics[width=1\linewidth]{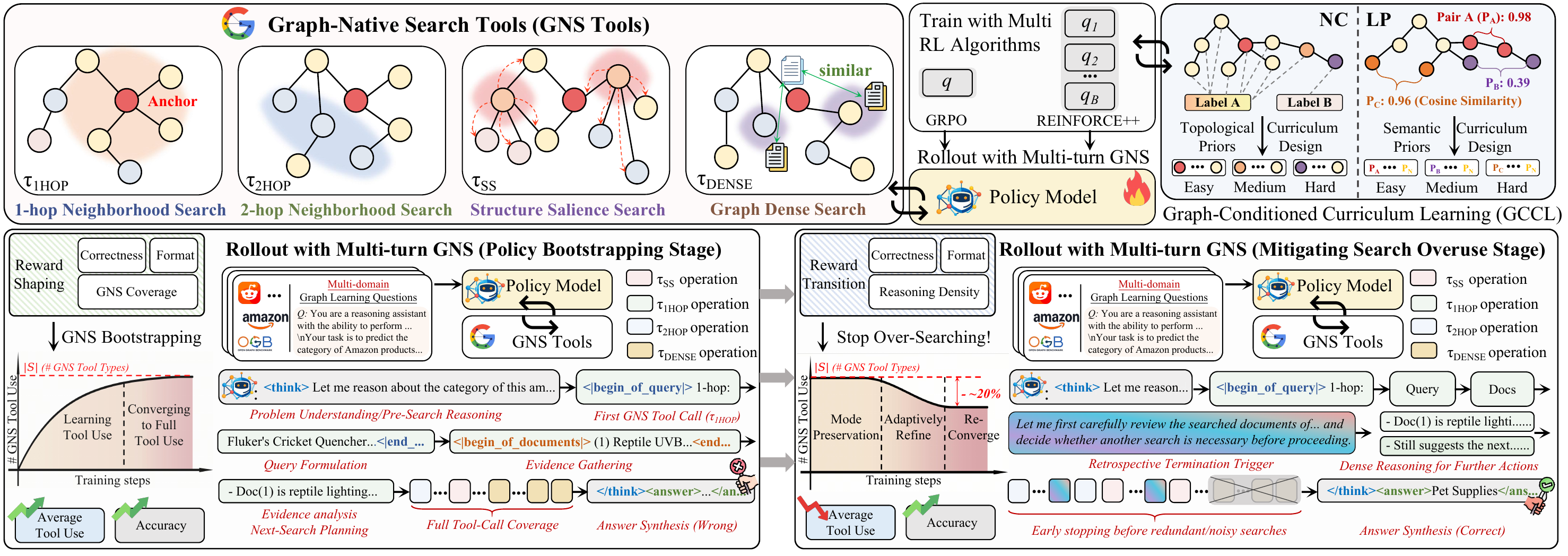}
    \caption{Method Overview. Equipped with graph-native search tools (Top-Left) for structural evidence mining, \texttt{AgentGL} employs a two-stage training strategy building on GCCL (Top-Right). The training progresses from Stage 1: Policy Bootstrapping (Bottom-Left), which uses shaped rewards to instill tool proficiency, to Stage 2: Mitigating Search Overuse (Bottom-Right), which optimizes the trade-off between search efficiency and reasoning accuracy.}
    \vspace{-12pt}
    \label{fig:agentgl}
\end{figure*}

We study agentic graph learning (AGL) on a TAG
\( \mathcal{G}=(\mathcal{V},\mathcal{A},\mathcal{T}) \),
where \( \mathcal{V} \) is the node set, \( \mathcal{A} \) is the adjacency matrix, and
\( \mathcal{T}=\{\mathbf{t}_v \mid v\in\mathcal{V}\} \) contains node texts.
In this paper, we focus on two classical GL tasks: Node Classification and Link Prediction.
Specifically, given a query $Q$, a target instance $x$ (e.g., a node \(v\in\mathcal{V}\) or a node pair \((u,v)\in\mathcal{V}\times\mathcal{V}\)) and ground-truth label \(y\),
the goal is to predict \(y\) by grounding the decision in graph-derived evidence.

\noindent\fbox{\parbox{0.465\textwidth}{
Formally, we formulate AGL as a sequential decision process on graph $\mathcal{G}$. 
Given a target $x$ and query $Q$, the policy $\pi_\theta$ iteratively samples actions $a_t \sim \pi_\theta(\cdot|h_t)$ from $\mathcal{S} \cup \{\textsc{Answer}\}$, where $\mathcal{S}$ denotes a predefined set of graph-native search tools for multi-scale exploration.
This interaction yields a trajectory $\tau$ containing the accumulated evidence $E$ and the final prediction $\hat{y}$. 
Our goal is to optimize $\theta$ to maximize the expected reward $\mathcal{R}$ over the dataset $\mathcal{D}$: $\mathcal{J}(\theta) = \mathbb{E}_{\tau \sim \pi_\theta} [\mathcal{R}]$.
}}

%% file: latex/05_methodology.tex
\section{Methodology}
% We present the \texttt{AgentGL} framework (Figure~\ref{fig:agentgl}), which employs a two-stage RL training pipeline: Graph-Native Search Policy Bootstrapping (Sec~\ref{subsec:stage1_bootstrap}) followed by Mitigating Search Overuse (Sec~\ref{sec:efficiency}). Both stages leverage Graph-Conditioned Curriculum Learning (Sec~\ref{subsec:gccl}) to enhance training stability and convergence speed.
We present the AgentGL framework (Figure \ref{fig:agentgl}), which organizes learning around two complementary components: graph-native policy bootstrapping (Sec. \ref{subsec:stage1_bootstrap}), where the agent acquires core navigation behaviors, and search-efficiency optimization (Sec. \ref{sec:efficiency}), which regulates tool use during long-horizon reasoning. Both components are trained under a graph-conditioned curriculum learning regime (Sec. \ref{subsec:gccl}) designed to improve stability and accelerate convergence.

\subsection{\texorpdfstring{{\fontsize{10.8}{13.5}\selectfont Graph-Native Search Policy Bootstrapping}}{Graph-Native Search Policy Bootstrapping}}
\label{subsec:stage1_bootstrap}
We begin by formulating the RL objective function, designed to empower the LLM agent to autonomously explore the graph structure while preserving its reasoning capabilities, defined as:
\[
\small
\mathcal{J}(\theta)
=
\mathbb{E}_{\substack{
(x,Q,y^*)\sim\mathcal{D}\\
\tau\sim\pi_\theta(\cdot\mid x,Q,\mathcal{G}_{\mathcal{S}})
}}
\Big[
\mathcal{R}(\hat{y},y^*)
-
\beta\cdot \mathbb{D}_{\text{KL}}(\pi_\theta \parallel \pi_{\text{ref}})
\Big].
\]
where $\mathcal{G}_{\mathcal{S}}$ signifies the graph environment accessed via the toolset $\mathcal{S} = \{\textcolor{InvBlue}{\tau_{\textsc{1hop}}}, \textcolor{FwdGreen}{\tau_{\textsc{2hop}}}, \textcolor{MutedPurple}{\tau_{\textsc{ss}}}, \textcolor{MutedRust}{\tau_{\textsc{dense}}}\}$; $\mathcal{R}(\hat{y}, y^*)$ is the outcome-based reward; $\mathbb{D}_{\text{KL}}$ represents the token-level KL divergence between the current policy $\pi_\theta$ and the reference policy $\pi_{\text{ref}}$; and $\beta$ is the coefficient controlling the KL penalty strength.
To bootstrap the graph-native search (GNS) policy, we next introduce the GNS tools in $\mathcal{S}$ for searching evidence (the text attributes of the collected candidates) directly from the TAG.

\begin{definition}[\textcolor{InvBlue}{\textit{\textbf{1-hop Neighborhood Search}}}]
\label{def:tool_1hop}
Given a query $Q$ and an input $x$, we simplify the process by treating $x$ as a pair $(u, v)$ (if $x$ is a single node $u$, we set $v=u$). Let $\mathcal{C} = \mathcal{N}_1(u) \cap \mathcal{N}_1(v)$ and $\mathcal{U}_z = \mathcal{N}_1(z) \setminus \mathcal{C}$ for $z \in \{u, v\}$.
The tool $\tau_{\textsc{1hop}}$ constructs the result set $E$ by prioritizing common neighbors and balancing exclusive ones:
\[
    E = \mathrm{TopK}(\mathcal{C}, K) \;\cup \bigcup_{z \in \{u, v\}} \mathrm{TopK}(\mathcal{U}_z, k_z),
\]

\noindent where the quotas $k_u, k_v$ satisfy $k_u + k_v = \max(0, K - |\mathcal{C}|) = R$ and represent the balanced allocation defined by:
\[
    k_u = \min\left(|\mathcal{U}_u|, \max\left(\lceil R/2 \rceil, R - |\mathcal{U}_v|\right)\right).
\]
\noindent Given the candidate nodes returned by the tools, the next question is how we select the most informative ones; we address this with a ranking score. The ranking score for a neighbor $n$ is computed via cosine similarity against a fusion embedding:
\[
    s(n) = \cos\left(\mathbf{h}_n, \; \lambda_r \mathbf{h}_Q + (1-\lambda_r) \mathbf{h}_x\right),
\]
where $\mathbf{h}_{(\cdot)}$ denotes the semantic embedding, $\mathbf{h}_x = \frac{1}{2}(\mathbf{h}_u + \mathbf{h}_v)$ averages the target pair, and $\lambda_r \in [0,1]$ balances the query relevance.
\end{definition}

\begin{definition}[\textcolor{FwdGreen}{\textit{\textbf{2-hop Neighborhood Search}}}]
\label{def:tool_2hop}
$\tau_{\textsc{2hop}}$ follows an analogous retrieval logic to Definition \ref{def:tool_1hop}, substituting scope $\mathcal{N}_1(\cdot)$ with  $\mathcal{N}_2(\cdot)$.
\end{definition}

\begin{definition}[\textcolor{MutedPurple}{\textit{\textbf{Structure Salience Search}}}]
\label{def:tool_salience}
Leveraging precomputed PPR scores \cite{jeh2003scaling} $s'(v)$, $\tau_{\textsc{ss}}$ retrieves the $\mathrm{TopK}$ globally salient candidates from the entire graph, ranking by $s'(v)$ for nodes or the mean $\frac{1}{2}(s(i)+s(j))$ for pairs.
\end{definition}

\begin{definition}[\textcolor{MutedRust}{\textit{\textbf{Graph Dense Search}}}]
\label{def:tool_dense}
The tool $\tau_{\textsc{dense}}$ operates identically to Definition \ref{def:tool_salience}, except that it substitutes the structural score $s'(\cdot)$ with the semantic relevance measured by the cosine similarity of node or pair embeddings $\phi(\cdot)$.
\end{definition}

\begin{remark}[Design Rationale]
\label{rem:rationale}
This toolkit is designed to ensure comprehensive coverage of the graph information space, spanning two critical dimensions: \textit{Local vs. Global} and \textit{Structure vs. Semantics}.  
$\tau_{\textsc{1hop}}$ and $\tau_{\textsc{2hop}}$ facilitate precise local grounding by harmonizing explicit topological dependencies with specific query requirements via a weighted fusion strategy.
To transcend structural isolation, the global operators offer broader horizons: $\tau_{\textsc{ss}}$ acts as a structural prior, identifying topological pivots to guide macro-level reasoning, while $\tau_{\textsc{dense}}$ adapts the Dense Retrieval paradigm from RAG to graphs, bridging disconnected nodes via latent semantic correlations. 
Collectively, these primitives empower LLMs to navigate the graph with the same versatility as navigating text.
\end{remark}

\noindent \textbf{Optimization with RL Algorithms.} To enable LLMs to leverage GNS tools for interleaved reasoning while exploring graph structure, and to avoid the high cost of constructing SFT-style supervision, we directly optimize the policy via RL. Specifically, we instantiate \texttt{AgentGL} with two mainstream critic-free policy optimization algorithms: Group Relative Policy Optimization (GRPO) \cite{shao2024deepseekmath} and REINFORCE++ (R++) \cite{hu2025reinforce} to optimize the AGL process. 

\noindent \textbf{\emph{Template Design and Trajectories.}}
To support interleaved reasoning and graph-native search in a way that is both learnable and automatically evaluable, we cast AGL as a \emph{reason--act--observe} interaction loop with a strict, machine-parseable interface.
Concretely, each prompt specifies (a) a dataset-routed task instruction with a closed label space and target instance text attribute, (b) a toolbox $\mathcal{S}$ of GNS pools with per-pool introduction, and (c) other instructions to steer the model toward the required response format.
Within the \tealtag{\texttt{<think>}}\textcolor{blue}{$\cdots$}\tealtag{\texttt{</think>}} block, the model may issue at most one retrieval action per round by emitting a pool-specific query tag
\searchtag{\texttt{<|begin\_of\_query|>}} \textcolor{blue}{\emph{tool name}\texttt{:}\emph{query}} \searchtag{\texttt{<|end\_of\_query|>}},
after which the environment executes the corresponding GNS tool and returns evidence wrapped in
\purpletag{\texttt{<|begin\_of\_documents|>}}\textcolor{blue}{$\cdots$}\purpletag{\texttt{<|end\_of\_documents|>}}.
Formally, let $h_0=(x,Q)$ denote the initial context.
We model the agentic rollout as a recursive state transition process, where the context evolves via the interactive trajectory defined by:
\[
\small
\label{eq:traj-fancy}
h_t = h_{t-1} \oplus \big( a_t,\llbracket a_t \rrbracket_{\mathcal{G}} \big) 
\quad \text{s.t.} \quad 
a_t \sim \pi_{\theta}(\cdot \mid h_{t-1})
\]
where the action $a_t=\langle s_t, q_t \rangle$ specifies a tool selector $s_t \in \mathcal{S}$ and a textual query $q_t$.
The semantic bracket $\llbracket a_t \rrbracket_{\mathcal{G}}$ denotes the structural evidence $o_t$ (e.g. text attributes) retrieved from graph $\mathcal{G}$.
The operator $\oplus$ recursively appends this interaction turn to the history $h_{t-1}$.
A rollout terminates either when the agent takes the terminal action and decides to output the final answer in
\rosetag{\texttt{<answer>}}\textcolor{blue}{$\cdots$}\rosetag{\texttt{</answer>}},
or when the maximum budget $B$ is exhausted.

\noindent \textbf{\emph{Reward Shaping.}}
We use a composite reward to provide dense, programmatic supervision for structured tool use while keeping the final objective aligned with task correctness.
Concretely, for a trajectory $\tau$ with prediction $\hat{y}$, we define
\[
R(\tau)= r_{\textsc{fmt}}(\tau)\;+\; r_{\textsc{acc}}(\hat{y},y)\;+\; r_{\textsc{cov}}(\tau)
\]
Format reward $r_{\textsc{fmt}}(\tau)$ enforces strict adherence to our tool-use template (tool name + query/args + structured think/answer), making trajectories reliably machine-parsable for stable RL.
Accuracy reward $r_{\textsc{acc}}(\hat{y},y)=\lambda_a\mathbb{I}[\hat{y}=y]$ anchors optimization to the end task and prevents reward hacking toward purely “well-formatted” behaviors.
GNS coverage reward $r_{\textsc{cov}}(\tau)$ encourages early exploration of all proposed tools, which is crucial to prevent early mode collapse to a single default action (one tool or no tool) and to ensure sufficient exploration over the discrete tool-action space. 
\[
\vspace{-2pt}
\mbox{\small $\displaystyle
r_{\textsc{cov}}(\tau)=\eta \sum_{j=1}^{\lvert \mathcal{S}\rvert}\mathbb{I}\!\left[\exists\,t:\ a_t=\tau_j\right],
\quad r_{\textsc{cov}}(\tau)\le \lvert \mathcal{S}\rvert \eta
$}
\vspace{-2pt}
\]

\noindent where each tool $\tau_j \in \mathcal{S}$ contributes at most once.

\subsection{Less is More: Mitigating Search Overuse}
\label{sec:efficiency}

While the bootstrapping stage establishes foundational graph navigation capabilities, it prioritizes \textit{feasibility} over \textit{optimality}, often defaulting to inefficient, exhaustive retrieval. However, given that the effective neighborhood range is highly instance-dependent \cite{xu2018representation}, the optimal structural context varies substantially across queries. Indiscriminate tool usage is thus counterproductive: it not only incurs computational overhead but creates structural noise that degrades reasoning fidelity. 
To address this, we introduce \textit{Search-Constrained Thinking}. This phase \textit{implicitly} optimizes efficiency by compelling the agent to autonomously discern the \textit{minimal sufficient trajectory}—maximizing accuracy by effectively pruning redundant steps. Accordingly, the optimization goal is formulated as:
\[
\vspace{-1pt}
\mbox{\small $\displaystyle
\theta^\star = \mathop{\mathrm{argmin}}_{\theta} \mathbb{E}_{\tau\sim\pi_\theta}[\,T(\tau)\,] \quad \text{s.t.} \quad \theta \in \operatorname*{argmax}_{\vartheta} \mathcal{J}_{\textsc{base}}(\vartheta)
$}
\vspace{-4pt}
\]
\noindent where $\mathcal{J}_{\textsc{base}}$ denotes the bootstrapping objective and $T(\tau)$ denotes the total search cost of trajectory $\tau$. By treating accuracy as a hard constraint, we restrict the efficiency optimization strictly to the optimal solution space, ensuring the agent learns parsimony without compromising performance.

\noindent \textbf{Search-Constrained Thinking.} 
To instantiate the implicit optimization target, we introduce a strategy that enforces a "Think more, Search less: Precision via Parsimony" paradigm. This approach couples retrospective verification with cognitive density constraints to substitute redundant retrieval with deep reasoning, via three components:

\noindent \textbf{\emph{Retrospective Termination Trigger.}} 
To preclude habitual search continuation, we inject a cognitive interrupt into the context after each tool execution:
\orangetag{Let me first carefully review the searched documents of \texttt{\{GNS tool name\}} and decide whether another search is necessary before proceeding.}
This trigger acts as a soft constraint, compelling the LLM to explicitly evaluate the sufficiency of the current evidence state $\mathcal{G}_\tau$ during training, transforming the search process from a habitual sequence into a series of deliberate, binary decisions.

\noindent \textbf{\emph{Cognitive Density Regularization.}} 
To ensure that reduced search frequency stems from efficient information absorption rather than superficial skipping, we impose a penalty on sparse reasoning. Formally, we define segments $\{s_i\}$ as the post-retrieval reasoning blocks dedicated to analyzing the acquired context and define a segment as ``deficient'' if the token length $\ell(s_i)$$<$ the threshold $\delta$. We introduce a depth-oriented term $r_{\text{depth}}$:
{\setlength{\abovedisplayskip}{3pt}
\setlength{\abovedisplayshortskip}{3pt}
\setlength{\belowdisplayskip}{3pt}
\setlength{\belowdisplayshortskip}{3pt}
\[
r_{\text{depth}}(z) = \alpha \cdot \mathbb{I}[N_{\text{short}} = 0] - \lambda_d \cdot N_{\text{short}}
\]}

\noindent where $N_{\text{short}}$ counts deficient segments. This formulation strictly penalizes fragmented thinking, incentivizing the generation of dense reasoning blocks before further actions.

\noindent \textbf{\emph{Adaptive Reward Transition.}} 
Reflecting the shift from exploration to exploitation, we discard the coverage incentive $r_{\textsc{cov}}$ while retaining $r_\textsc{fmt}$ for format constraints. The main optimization is thus streamlined to the synergistic maximization of accuracy $r_\textsc{acc}$ and reasoning density $r_{{depth}}$. This alignment prioritizes deep internal processing over redundant retrieval, naturally converging onto the minimal sufficient trajectory:
\[
R(\tau)= r_{\textsc{fmt}}(\tau)\;+\; r_{\textsc{acc}}(\hat{y},y)\;+\; r_{depth}(z)
\]

\subsection{Graph-Conditioned    Curriculum Learning}
\label{subsec:gccl}
To stabilize training and accelerate convergence, we leverage intrinsic graph properties for curriculum design. Unlike reasoning tasks where difficulty estimation relies on expert annotation \cite{hendrycks2021measuring} or expensive pilot rollouts \cite{song2025r1}, graphs offer a distinct advantage: learnability is directly quantifiable via topological and semantic priors. We formulate an analytical difficulty scoring function $\mathcal{S}(\cdot)$ to proxy hardness, enabling a smooth, cost-free training progression from confident to ambiguous instances for different tasks via graph-conditioned curriculum learning (GCCL).
\paragraph{Node Classification with GCCL.} 
Drawing on prior theoretical insights, node classification difficulty is jointly governed by local homophily and degree magnitude \cite{tang2020investigating, zhu2020beyond} in many cases. To derive a robust metric $\mathcal{S}_{\text{NC}}(v)$ that approximately estimates difficulty, we rectify homophily estimates using the Wilson Lower Bound, augmented by degree magnitude:
\[
\vspace{-1pt}
\resizebox{0.94\linewidth}{!}{$
\mathcal{S}_{\text{NC}}(v) =
\underbrace{\frac{\hat{p}_v + \frac{z^2}{2d_v} - z\sqrt{\frac{\hat{p}_v(1-\hat{p}_v)}{d_v} + \frac{z^2}{4d_v^2}}}{1 + \frac{z^2}{d_v}}}_{\text{Wilson Lower Bound}}
\;+\; \eta \log(1+d_v)
$}
\vspace{-1pt}
\]
\noindent where $\hat{p}_v$ is the neighbor label consistency, $d_v$ is the degree, $z$ is the standard normal quantile and $\eta$ regulates the impact of degree priors. This formulation prioritizes structurally prominent hubs (\textit{Easy}), progresses through intermediate nodes (\textit{Medium}), and defers ambiguous, heterophilous outliers (\textit{Hard}).

\paragraph{Link Prediction with GCCL.} 
Inspired by heuristics in link prediction \cite{zhang2018link, mao2023revisiting}, we posit that ``easiness'' aligns with the consistency between semantic similarity and label existence. For a link pair $e=(u,v)$ with label $y_e \in \{0,1\}$, we calculate the score based on cosine similarity of node features $\text{sim}(\mathbf{x}_u, \mathbf{x}_v)$:
\[
\vspace{-1pt}
\mbox{\small $\displaystyle
\mathcal{S}_{\text{LP}}(e)
= y_e \cdot \text{sim}(\mathbf{x}_u, \mathbf{x}_v)
+ (1 - y_e)\cdot\bigl(1-\text{sim}(\mathbf{x}_u, \mathbf{x}_v)\bigr)
$}
\vspace{-1pt}
\]

\noindent We prioritize consistent pairs (high-sim positives, low-sim negatives) as \textit{Easy}. The curriculum traverses ambiguous \textit{Medium} instances, deferring \textit{Hard} structural noise-conflicting cases like high-sim negatives to later training iterations.

\noindent \textbf{Training Process.} 
Algorithm \ref{alg:agentgl_training_short} outlines the procedure: AgentGL first undergoes graph-native policy bootstrapping (Sec. \ref{subsec:stage1_bootstrap}), then search-efficiency refinement (Sec. \ref{sec:efficiency}). Both stages follow an easy-to-hard curriculum in Sec. \ref{subsec:gccl}. More details are provided in Appendix \ref{app:implementation}.
% In both stages, training follows an easy-to-hard curriculum within the corresponding data split (as noted in the Algorithm~\ref{alg:agentgl_training_short}).

%% file: latex/06_experiments.tex
\section{Experiments}

\begin{table*}[t]\scriptsize
\centering
\scriptsize{
\resizebox{\linewidth}{!}{
\setlength\tabcolsep{2pt}
\renewcommand\arraystretch{1.1}
\setlength{\arrayrulewidth}{0.3mm}
\begin{tabular}{c||cc|ccccc||cc|ccccc}
\hline

% -------------------- Header Row 1 --------------------
\rowcolor{HdrA}
\textbf{Tasks}
& \multicolumn{7}{c||}{\textbf{Node Classification}}
& \multicolumn{7}{c}{\textbf{Link Prediction}}\\
\hline
% -------------------- Header Row 2 --------------------
\rowcolor{HdrB}
\textbf{Settings}
& \multicolumn{2}{c|}{\textbf{In-Domain}}
& \multicolumn{5}{c||}{\textbf{Zero-shot Transfer}}
& \multicolumn{2}{c|}{\textbf{In-Domain}}
& \multicolumn{5}{c}{\textbf{Zero-shot Transfer}}\\
\hline
% -------------------- Header Row 3 --------------------
\rowcolor{HdrC}
\multirow{-1}{*}{\centering \textbf{Methods}}
& \textbf{OGB-Arxiv} & \textbf{OGB-Products} & \textbf{PubMed} & \textbf{Photo} & \textbf{Computers} & \textbf{Arxiv-23} & \textbf{Reddit}
& \textbf{OGB-Arxiv} & \textbf{OGB-Products} & \textbf{PubMed} & \textbf{Photo} & \textbf{Computers} & \textbf{Arxiv-23} & \textbf{Reddit}\\
\hline
\hline

% ==================== GNN Baselines ====================
\rowcolor{HdrD}
\multicolumn{15}{c}{\textbf{GNN-Based Methods}}\\
\hline
\hline
\rowcolor{gray!3}
GCN\,\scalebox{0.85}{\faCodeBranch}
& $60.2$ \myred{7.7} & $58.8$ \myred{10.8} & $14.1$ \myred{64.5} & $8.0$ \myred{42.8} & $10.0$ \myred{50.2} & $1.9$ \myred{63.6} & $7.1$ \myred{40.0}
& $55.5$ \myred{38.1} & $74.7$ \myred{18.8} & $51.0$ \myred{24.8} & $49.8$ \myred{19.7} & $50.5$ \myred{24.1} & $52.0$ \myred{38.1} & $53.2$ \myred{34.8}\\

RevGAT\,\scalebox{0.85}{\faCodeBranch}
& $58.9$ \myred{9.0} & $56.7$ \myred{12.9} & $13.5$ \myred{65.1} & $12.1$ \myred{38.8} & $5.6$ \myred{54.6} & $1.3$ \myred{64.2} & $8.2$ \myred{38.9}
& $59.0$ \myred{34.6} & $73.5$ \myred{20.0} & $55.8$ \myred{20.0} & $59.4$ \myred{10.1} & $50.7$ \myred{23.9} & $61.8$ \myred{28.3} & $61.8$ \myred{26.2}\\

\rowcolor{gray!3}
SAGE\,\scalebox{0.85}{\faCodeBranch}
& $60.1$ \myred{7.8} & $56.9$ \myred{12.7} & $11.1$ \myred{67.5} & $8.2$ \myred{42.7} & $7.4$ \myred{52.8} & $3.6$ \myred{61.9} & $3.1$ \myred{44.0}
& $61.0$ \myred{32.6} & $73.4$ \myred{20.1} & $50.9$ \myred{24.9} & $51.6$ \myred{17.9} & $50.3$ \myred{24.3} & $50.5$ \myred{39.6} & $57.6$ \myred{30.4}\\
\hline
\hline

% ==================== LLM Baselines (Qwen2.5 3B) ====================
\rowcolor{HdrD}
\multicolumn{15}{c}{\textbf{LLM-based Methods}}\\
\hline
\hline
\rowcolor{HdrE}
\multicolumn{15}{c}{\textbf{Qwen2.5-3B-Instruct}}\\
\hline
GraphPrompter\,\scalebox{0.85}{\faPuzzlePiece} & $54.1$ \myred{12.2} & $61.0$ \myred{1.3} & $67.0$ \myred{7.5} & $8.5$ \myred{34.2} & $29.2$ \myred{23.1} & $37.9$ \myred{25.7} & $4.7$ \myred{35.0} & $79.6$ \myred{11.8} & $74.4$ \myred{15.7} & $60.9$ \myred{10.9} & $50.9$ \myred{14.9} & $52.7$ \myred{15.1} & $55.8$ \myred{32.1} & $46.8$ \myred{36.4}\\
\rowcolor{gray!3}
GraphGPT\,\scalebox{0.85}{\faPuzzlePiece} & $12.5$ \myred{53.8} & $23.4$ \myred{38.9} & $62.3$ \myred{12.2} & $3.0$ \myred{39.7} & $11.5$ \myred{40.8} & $44.6$ \myred{19.0} & $9.5$ \myred{30.2} & $59.9$ \myred{31.6} & $43.3$ \myred{46.8} & $44.5$ \myred{27.4} & $54.5$ \myred{11.3} & $51.1$ \myred{16.6} & $63.4$ \myred{24.5} & $49.8$ \myred{33.4}\\
LLaGA\,\scalebox{0.85}{\faPuzzlePiece} & $54.6$ \myred{11.6} & $57.9$ \myred{4.4} & $61.8$ \myred{12.7} & $4.2$ \myred{38.5} & $12.3$ \myred{40.0} & $44.8$ \myred{18.8} & $10.1$ \myred{29.6} & $79.8$ \myred{11.7} & $73.6$ \myred{16.5} & $61.5$ \myred{10.4} & $51.3$ \myred{14.5} & $53.2$ \myred{14.6} & $52.1$ \myred{35.8} & $43.7$ \myred{39.5}\\
\rowcolor{gray!3}
GraphICL\,\scalebox{0.85}{\faPuzzlePiece} & $49.4$ \myred{16.9} & $45.8$ \myred{16.5} & $53.8$ \myred{20.7} & $35.2$ \myred{7.5} & $42.4$ \myred{9.9} & $52.2$ \myred{11.4} & $37.1$ \myred{2.6} & $76.3$ \myred{15.2} & $69.2$ \myred{20.9} & $62.5$ \myred{9.4} & $54.8$ \myred{11.0} & $57.7$ \myred{10.1} & $57.6$ \myred{30.3} & $62.0$ \myred{21.2}\\
\rowcolor{gray!3}
LinearRAG\,\scalebox{0.85}{\faProjectDiagram} & $43.8$ \myred{22.4} & $52.5$ \myred{9.8} & $64.2$ \myred{10.3} & $11.8$ \myred{30.9} & $20.1$ \myred{32.2} & $28.9$ \myred{34.7} & $32.7$ \myred{7.0} & $47.3$ \myred{44.1} & $45.3$ \myred{44.8} & $46.0$ \myred{25.9} & $47.7$ \myred{18.1} & $45.7$ \myred{22.0} & $45.9$ \myred{42.0} & $50.1$ \myred{33.1}\\
HippoRAG2\,\scalebox{0.85}{\faProjectDiagram} & $44.2$ \myred{22.0} & $53.2$ \myred{9.1} & $63.5$ \myred{11.0} & $12.2$ \myred{30.4} & $23.8$ \myred{28.5} & $30.6$ \myred{32.9} & $34.7$ \myred{5.0} & $48.1$ \myred{43.4} & $46.7$ \myred{43.4} & $45.9$ \myred{26.0} & $48.1$ \myred{17.7} & $45.8$ \myred{21.9} & $48.5$ \myred{39.4} & $49.7$ \myred{33.5}\\
\rowcolor{gray!3}
GraphCoT\,\scalebox{0.85}{\faProjectDiagram}/\scalebox{0.85}{\faAndroid} & $44.3$ \myred{22.0} & $56.1$ \myred{6.1} & $72.8$ \myred{1.7} & $38.0$ \myred{4.7} & $47.6$ \myred{4.7} & $40.1$ \myred{23.5} & $37.2$ \myred{2.5} & $50.9$ \myred{40.6} & $51.0$ \myred{39.1} & $52.0$ \myred{19.9} & $50.7$ \myred{15.1} & $50.2$ \myred{17.6} & $51.1$ \myred{36.8} & $50.3$ \myred{32.8}\\
Qwen2.5\,\scalebox{0.85}{\faRobot} & $49.1$ \myred{17.2} & $54.2$ \myred{8.1} & $57.7$ \myred{16.8} & $10.3$ \myred{32.4} & $14.5$ \myred{37.8} & $40.8$ \myred{22.8} & $9.9$ \myred{29.8} & $72.2$ \myred{19.3} & $70.3$ \myred{19.8} & $54.9$ \myred{16.9} & $51.3$ \myred{14.5} & $51.5$ \myred{16.3} & $48.9$ \myred{39.0} & $43.2$ \myred{40.0}\\
\rowcolor{gray!3}
Search-R1\,\scalebox{0.85}{\faSearch} & $60.2$ \myred{6.1} & $58.3$ \myred{4.0} & $71.4$ \myred{3.1} & $37.8$ \myred{4.9} & $42.7$ \myred{9.6} & $55.9$ \myred{7.6} & $38.7$ \myred{1.0} & $82.5$ \myred{8.9} & $84.9$ \myred{5.2} & $61.8$ \myred{10.1} & $54.6$ \myred{11.2} & $62.2$ \myred{5.6} & $79.6$ \myred{8.3} & $69.0$ \myred{14.2}\\
Search-O1\,\scalebox{0.85}{\faSearch} & $48.2$ \myred{18.1} & $44.7$ \myred{17.6} & $53.9$ \myred{20.6} & $34.6$ \myred{8.1} & $40.1$ \myred{12.2} & $51.6$ \myred{11.9} & $38.2$ \myred{1.5} & $74.4$ \myred{17.1} & $65.8$ \myred{24.3} & $59.7$ \myred{12.1} & $54.9$ \myred{10.9} & $56.9$ \myred{10.9} & $73.2$ \myred{14.7} & $54.5$ \myred{28.7}\\
\hline

% ==================== Ours (highlight in yellow) ====================
\rowcolor[HTML]{FFF0C1}
\textbf{AgentGL-R++}\,\scalebox{0.85}{\faChessKing} & \textbf{$65.6$} & $\mathbf{63.3}$ & \textbf{$73.6$} & $\mathbf{42.7}$ & \textbf{$49.8$} & \textbf{$60.2$} & $\mathbf{40.2}$ & $\mathbf{92.3}$ & $\mathbf{92.6}$ & $\mathbf{72.3}$ & \textbf{$65.2$} & $\mathbf{68.7}$ & $\mathbf{90.3}$ & $\mathbf{86.6}$\\

\rowcolor[HTML]{FFF0C1}
\textbf{AgentGL-GRPO}\,\scalebox{0.85}{\faChessKing} & $\mathbf{66.9}$ & \textbf{$61.2$} & $\mathbf{75.4}$ & \textbf{$42.6$} & $\mathbf{54.8}$ & $\mathbf{66.9}$ & \textbf{$39.2$} & \textbf{$90.6$} & \textbf{$87.6$} & \textbf{$71.4$} & $\mathbf{66.4}$ & \textbf{$66.8$} & \textbf{$85.4$} & \textbf{$79.7$}\\
\hline

% ==================== LLM Baselines (Qwen2.5 7B) ====================
\rowcolor{HdrE}
\multicolumn{15}{c}{\textbf{Qwen2.5-7B-Instruct}}\\
\hline
GraphPrompter\,\scalebox{0.85}{\faPuzzlePiece} & $60.8$ \myred{8.8} & $70.2$ \myred{6.7} & $80.6$ \myred{2.1} & $39.8$ \myred{19.3} & $45.2$ \myred{22.9} & $62.8$ \myred{4.7} & $24.9$ \myred{29.6} & $83.2$ \myred{12.6} & $80.8$ \myred{16.1} & $66.9$ \myred{12.9} & $56.1$ \myred{17.1} & $61.2$ \myred{20.3} & $56.8$ \myred{35.6} & $50.9$ \myred{41.9}\\
\rowcolor{gray!3}
GraphGPT\,\scalebox{0.85}{\faPuzzlePiece} & $53.8$ \myred{15.8} & $59.1$ \myred{17.8} & $79.2$ \myred{3.5} & $4.4$ \myred{54.7} & $12.3$ \myred{55.8} & $66.0$ \myred{1.5} & $32.2$ \myred{22.3} & $50.5$ \myred{45.3} & $48.2$ \myred{48.8} & $48.7$ \myred{31.1} & $49.8$ \myred{23.4} & $44.4$ \myred{37.1} & $54.7$ \myred{37.7} & $33.2$ \myred{59.6}\\
LLaGA\,\scalebox{0.85}{\faPuzzlePiece} & $62.8$ \myred{6.8} & $67.4$ \myred{9.5} & $78.6$ \myred{4.1} & $5.6$ \myred{53.4} & $14.6$ \myred{53.4} & $67.2$ \myred{0.3} & $31.7$ \myred{22.8} & $84.1$ \myred{11.7} & $79.3$ \myred{17.7} & $67.7$ \myred{12.1} & $56.8$ \myred{16.4} & $60.2$ \myred{21.3} & $56.7$ \myred{35.7} & $48.5$ \myred{44.3}\\
\rowcolor{gray!3}
GraphICL\,\scalebox{0.85}{\faPuzzlePiece} & $66.4$ \myred{3.2} & $61.3$ \myred{15.6} & $70.4$ \myred{12.2} & $42.1$ \myred{17.0} & $49.0$ \myred{19.1} & $51.0$ \myred{16.5} & $48.8$ \myred{5.6} & $78.2$ \myred{17.6} & $80.9$ \myred{16.1} & $72.3$ \myred{7.5} & $57.6$ \myred{15.6} & $61.8$ \myred{19.7} & $72.4$ \myred{20.0} & $63.3$ \myred{29.5}\\
\rowcolor{gray!3}
LinearRAG\,\scalebox{0.85}{\faProjectDiagram} & $48.2$ \myred{21.4} & $58.3$ \myred{18.6} & $72.7$ \myred{10.0} & $47.6$ \myred{11.5} & $58.3$ \myred{9.8} & $53.4$ \myred{14.1} & $51.4$ \myred{3.1} & $47.6$ \myred{48.2} & $48.2$ \myred{48.8} & $47.2$ \myred{32.6} & $46.3$ \myred{26.9} & $46.8$ \myred{34.7} & $47.5$ \myred{44.9} & $49.8$ \myred{43.0}\\
HippoRAG2\,\scalebox{0.85}{\faProjectDiagram} & $49.6$ \myred{20.0} & $57.2$ \myred{19.7} & $75.3$ \myred{7.4} & $52.0$ \myred{7.1} & $59.0$ \myred{9.1} & $51.0$ \myred{16.5} & $50.2$ \myred{4.2} & $48.3$ \myred{47.5} & $47.1$ \myred{49.9} & $46.3$ \myred{33.5} & $48.2$ \myred{25.0} & $44.5$ \myred{37.0} & $47.6$ \myred{44.8} & $50.5$ \myred{42.3}\\
\rowcolor{gray!3}
GraphCoT\,\scalebox{0.85}{\faProjectDiagram}/\scalebox{0.85}{\faAndroid} & $53.1$ \myred{16.5} & $62.6$ \myred{14.3} & $81.9$ \myred{0.8} & $44.2$ \myred{14.9} & $54.4$ \myred{13.7} & $51.6$ \myred{15.9} & $49.8$ \myred{4.7} & $50.4$ \myred{45.3} & $52.4$ \myred{44.6} & $50.5$ \myred{29.2} & $51.0$ \myred{22.2} & $50.2$ \myred{31.3} & $49.9$ \myred{42.5} & $52.4$ \myred{40.4}\\
Qwen2.5\,\scalebox{0.85}{\faRobot} & $54.7$ \myred{14.9} & $55.0$ \myred{21.9} & $63.8$ \myred{18.9} & $16.4$ \myred{42.7} & $29.5$ \myred{38.6} & $60.0$ \myred{7.5} & $30.9$ \myred{23.6} & $74.4$ \myred{21.3} & $76.3$ \myred{20.7} & $60.2$ \myred{19.5} & $53.1$ \myred{20.1} & $46.5$ \myred{35.0} & $50.9$ \myred{41.5} & $47.1$ \myred{45.7}\\
\rowcolor{gray!3}
Search-R1\,\scalebox{0.85}{\faSearch} & $63.2$ \myred{6.4} & $70.4$ \myred{6.5} & $81.6$ \myred{1.1} & $46.2$ \myred{12.9} & $54.6$ \myred{13.5} & $66.8$ \myred{0.7} & $50.7$ \myred{3.7} & $86.6$ \myred{9.2} & $89.1$ \myred{7.8} & $72.3$ \myred{7.5} & $62.1$ \myred{11.1} & $70.1$ \myred{11.4} & $80.2$ \myred{12.2} & $74.1$ \myred{18.7}\\
Search-O1\,\scalebox{0.85}{\faSearch} & $60.2$ \myred{9.4} & $59.9$ \myred{17.0} & $69.5$ \myred{13.2} & $41.7$ \myred{17.4} & $49.3$ \myred{18.8} & $53.8$ \myred{13.7} & $47.2$ \myred{7.2} & $75.1$ \myred{20.6} & $73.9$ \myred{23.1} & $65.8$ \myred{13.9} & $57.6$ \myred{15.6} & $60.6$ \myred{20.9} & $74.0$ \myred{18.4} & $59.0$ \myred{33.8}\\
\hline
% ==================== Ours again (highlight in yellow) ====================
\rowcolor[HTML]{FFF0C1}
\textbf{AgentGL-R++}\,\scalebox{0.85}{\faChessKing} & $\mathbf{70.3}$ & \textbf{$76.8$} & \textbf{$82.6$} & \textbf{$58.2$} & \textbf{$67.5$} & $\mathbf{67.6}$ & $\mathbf{54.8}$ & \textbf{$95.6$} & $\mathbf{97.4}$ & $\mathbf{80.4}$ & $\mathbf{77.0}$ & $\mathbf{87.3}$ & \textbf{$90.5$} & $\mathbf{97.1}$\\

\rowcolor[HTML]{FFF0C1}
\textbf{AgentGL-GRPO}\,\scalebox{0.85}{\faChessKing} & \textbf{$68.9$} & $\mathbf{77.0}$ & $\mathbf{82.7}$ & $\mathbf{59.9}$ & $\mathbf{68.6}$ & \textbf{$67.4$} & \textbf{$54.1$} & $\mathbf{95.9}$ & \textbf{$96.5$} & \textbf{$79.1$} & \textbf{$69.4$} & \textbf{$75.7$} & $\mathbf{94.3}$ & \textbf{$88.5$}\\
\hline

\end{tabular}}}
\caption{Performance Comparison on Node Classification and Link Prediction benchmarks under In-Domain and Zero-shot Transfer settings. The best results are highlighted in bold. The metric used for two tasks is ACC (\%). \textcolor{red}{Red} values ($\uparrow$) indicate the absolute gain over baselines. These denote the average improvement of the two RL variants; for GNN comparisons, they represent the average gain across 3B and 7B backbones.}
\label{tab:Main Results}
\end{table*}

\begin{table*}[t]\scriptsize
\vspace{-5pt}
\centering
\scriptsize{
\resizebox{\linewidth}{!}{%
    \setlength\tabcolsep{6pt}
    \renewcommand\arraystretch{1.2}
    \setlength{\arrayrulewidth}{0.3mm}
    \begin{tabular}{c|c||cc|cc|cc|cc|cc|cc|cc}
        \hline \thickhline
        \rowcolor{CadetBlue!20}
        && \multicolumn{14}{c}{Datasets} \\
        \cline{3-16}
        \rowcolor{CadetBlue!20}
        \multirow{-2}{*}{\textsc{GNSPB}} & \multirow{-2}{*}{\textsc{MSO}}
        & \multicolumn{2}{c|}{\textsc{OGB-Arxiv}}
        & \multicolumn{2}{c|}{\textsc{OGB-Products}}
        & \multicolumn{2}{c|}{\textsc{Arxiv-2023}}
        & \multicolumn{2}{c|}{\textsc{PubMed}}
        & \multicolumn{2}{c|}{\textsc{Amazon-Photo}}
        & \multicolumn{2}{c|}{\textsc{Amazon-Computers}}
        & \multicolumn{2}{c}{\textsc{Reddit}} \\
        \rowcolor{CadetBlue!20}
        && \textsc{ACC}(\%) & \#Search
        & \textsc{ACC}(\%) & \#Search
        & \textsc{ACC}(\%) & \#Search
        & \textsc{ACC}(\%) & \#Search
        & \textsc{ACC}(\%) & \#Search
        & \textsc{ACC}(\%) & \#Search
        & \textsc{ACC}(\%) & \#Search \\
        \hline\hline

        \ding{51} & \ding{55}
        & $65.6$ \myred{3.3} & $4.00$ \blue{18.8\%}
        & $74.1$ \myred{2.9} & $4.00$ \blue{14.0\%}
        & $66.2$ \myred{1.2} & $3.98$ \blue{10.5\%}
        & $80.8$ \myred{1.9} & $3.99$ \blue{26.0\%}
        & $56.9$ \myred{3.0} & $3.99$ \blue{20.5\%}
        & $67.2$ \myred{1.4} & $3.97$ \blue{19.0\%}
        & $50.7$ \myred{3.4} & $3.97$ \blue{13.5\%} \\

        \ding{55} & \ding{51}
        & $62.3$ \myred{6.6} & $0.00$ \myred{81.2\%}
        & $71.9$ \myred{5.1} & $0.00$ \myred{86.0\%}
        & $61.8$ \myred{5.6} & $0.11$ \myred{86.2\%}
        & $79.1$ \myred{3.6} & $0.02$ \myred{73.2\%}
        & $54.3$ \myred{5.6} & $0.01$ \myred{79.0\%}
        & $60.7$ \myred{7.9} & $0.01$ \myred{80.0\%}
        & $48.7$ \myred{5.4} & $0.02$ \myred{85.2\%} \\

        \rowcolor[HTML]{D7F6FF}
        \ding{51} & \ding{51}
        & $68.9$ & $3.25$
        & $77.0$ & $3.44$
        & $67.4$ & $3.56$
        & $82.7$ & $2.95$
        & $59.9$ & $3.17$
        & $68.6$ & $3.21$
        & $54.1$ & $3.43$ \\

        \hline
    \end{tabular}%
}}
\vspace{-3pt}
\caption{Ablation study on different RL training stages. The \textcolor{red}{red} ($\uparrow$) and \textcolor{green}{green} ($\downarrow$) denote absolute percentage gains and declines, respectively. \#Search represents the average search count on the test set of each dataset (Budget=4). The percentage calculation and comparison for the number of searches are conducted based on the budget.}
\label{tab:ablation_rl_stage}
\vspace{-12pt}
\end{table*}

We conduct extensive experiments to validate the effectiveness of \texttt{AgentGL}. Specifically, we evaluate on 7 TAG datasets spanning 3 domains, compare against 13 baselines across five categories, and test 2 backbone LLMs with different parameter scales. 
% See the Appendix for details.

\noindent \textbf{Datasets.} We use the following datasets: (1) \emph{\textbf{Citation Networks}}: OGB-Arxiv \cite{hu2020open}, PubMed \cite{sen2008collective}, and Arxiv-2023 \cite{he2023harnessing}; (2) \emph{\textbf{Amazon Products}}: OGB-Products \cite{hu2020open}, Amazon-Photo, and Amazon-Computers \cite{shchur2018pitfalls}; and (3) \emph{\textbf{Social Networks}}: Reddit \cite{yan2025graph}. 
% In these TAGs, each node corresponds to a paper, a product, or a post, and edges encode citation, co-purchase, or co-posting relations. 
Additional dataset details and data splits are provided in the Appendix~\ref{app:datasets}. 

\noindent \textbf{Baselines.} We compare \texttt{AgentGL} against a diverse set of up-to-date, strong baselines spanning (1) \emph{\textbf{GNNs}} (\scalebox{0.85}{\faCodeBranch}): GraphSAGE (\citeyear{li2021training}), GCN (\citeyear{kipf2016semi}) and RevGAT (\citeyear{li2021training}); (2) \emph{\textbf{GraphLLMs} (\scalebox{0.85}{\faPuzzlePiece})}: LLaGA  (\citeyear{chen2024llaga}), GraphGPT (\citeyear{tang2023graphgpt}), GraphPrompter (\citeyear{liu2024can}) and GraphICL(-S1) (\citeyear{sun2025graphicl}); (3) \emph{\textbf{GraphRAG} (\scalebox{0.85}{\faProjectDiagram})}: LinearRAG (\citeyear{zhuang2025linearrag}), HippoRAG2 (\citeyear{gutierrez2025rag}) and GraphCoT (A special kind of agent framework \scalebox{0.85}{\faAndroid}) (\citeyear{jin2024graph}); (4) \emph{\textbf{Standard Agentic Search} (\scalebox{0.85}{\faSearch})}: Search-R1 (\citeyear{jin2025search}) and Search-O1 (\citeyear{li2025search}); (5) \emph{\textbf{Large Language Models (\scalebox{0.85}{\faRobot})}}: Qwen2.5-3B/7B-Instruct (\citeyear{qwen2025qwen25}) (SFT), to comprehensively assess predictive performance.

\noindent \textbf{Setup.} For a fair comparison, we use the same LLM backbone as \texttt{AgentGL} for all baselines whose final reasoner is an LLM. For GraphRAG baselines, we construct the graph-based retrieval corpus by collecting the node texts involved in our experiments. For standard agentic search baselines, since they lack native graph-search capability, we replace their original online-search space with the set of graph nodes, ensuring that they can be properly applied to graph reasoning tasks; implementation details are provided in the appendix. Other settings, such as SFT and RL training procedures, follow the original papers. The other implementation details are provided in the Appendix~\ref{app:implementation}.

\subsection{Overall Performance}

We first evaluate in-domain and zero-shot transfer performance, with results listed in Table~\ref{tab:Main Results}. For all methods, we train only on OGB-Arxiv and OGB-Products using different Qwen backbones, and then test on the test splits of all datasets. Based on these results, we summarize the following observations:

\noindent \textbf{\textit{Obs 1. AgentGL consistently achieves the best performance across multiple tasks and domains, under diverse graph-reasoning regimes.}} 
For node classification (NC), with Qwen7B as the backbone, \texttt{AgentGL} outperforms the baselines by an average of \textbf{12.7\%} on the in-domain evaluation and \textbf{24.4\%} on the zero-shot transfer setting. 
For link prediction (LP), \texttt{AgentGL} achieves an average gain of \textbf{26.3\%} in-domain and \textbf{22.4\%} in zero-shot transfer. 
These improvements are consistent across model scales: with Qwen3B, \texttt{AgentGL} improves over the baselines by \textbf{14.5\%} on in-domain NC and \textbf{26.3\%} on in-domain LP, and by \textbf{26.6\%} and \textbf{22.4\%} on zero-shot NC and LP, respectively.

\noindent \textbf{\textit{Obs 2. AgentGL showcases the promise of interleaved graph reasoning and searching, outperforming static context stuffing.}} Methods that primarily rely on static “stuffing” (e.g., GraphRAG or GraphLLM) can be competitive in some settings, but are consistently outperformed by \texttt{AgentGL}. Taking Qwen7B LP as an example, \texttt{AgentGL} achieves \textbf{47.4\%} and \textbf{23.2\%} higher in-domain performance than GraphRAG and GraphLLM, respectively; these substantial margins are sustained under zero-shot transfer at \textbf{35.4\%} and \textbf{26.9\%}. This trend suggests that static context injection is more brittle to distribution shifts, while \texttt{AgentGL}’s interleaved searching, reasoning loop can adaptively acquire task-relevant evidence and suppress irrelevant context, leading to more robust transfer.

\noindent \textbf{\textit{Obs 3. Different RL algorithms yield complementary strengths for AgentGL across graph tasks.}} Across datasets, \texttt{AgentGL}-R++ and \texttt{AgentGL}-GRPO show a consistent stage/algorithm-dependent profile: GRPO yields higher NC performance by an average of \textbf{0.9\%} across settings (averaged over Qwen3B/7B), whereas R++ is stronger on LP, improving over GRPO by \textbf{3.3\%} on average across settings. This indicates a clear tradeoff between algorithms, with task-wise advantages that can be selected based on the target domain.

\noindent \textbf{\textit{Obs 4. Scaling up the backbone enhances AgentGL’s agentic graph learning capability.}} 
Scaling the backbone from 3B to 7B consistently improves \texttt{AgentGL} on both tasks: the average gain is \textbf{9.0\%} (in-domain) and \textbf{11.8\%} (zero-shot) for NC, and \textbf{5.6\%} (in-domain) and \textbf{8.7\%} (zero-shot) for LP. The improvement is particularly pronounced under zero-shot transfer, indicating that larger backbones better learn and generalize the tool-use policy for adaptive evidence acquisition.

\subsection{Impact of Multi-Stage Training}
To study the impact of our proposed two-stage RL training: GNS Policy Bootstrapping (GNSPB) and Mitigating Search Overuse (MSO), on both overall performance and search efficiency, we conduct a stage-wise ablation analysis. Specifically, we ablate each training stage and compare the resulting variants in terms of accuracy and tool-call cost, with results reported in Table~\ref{tab:ablation_rl_stage}.

\noindent \textbf{\textit{Obs 5. Omitting any RL training stage of AgentGL leads to concurrent drops in both efficiency and performance.}} When keeping only the \textsc{GNSPB} stage, the presence of $r_{\textsc{cov}}(\tau)$ encourages the LLM to reliably cover all four tools, which maintains relatively strong performance across datasets; however, it almost always consumes near-full search budgets, increasing the overall cost. In contrast, when keeping only the \textsc{MSO} stage, the policy tends to collapse during training, converging to the degenerate behavior of making zero searches and carrying this pattern over to inference, which weakens its search capability and degrades performance, resulting in the worst overall results. Only by combining both stages can the model achieve both strong performance and high efficiency; for example, compared to using \textsc{GNSPB} alone, the full method reduces tool calls by about \textbf{17.5\%} while improving accuracy by an average of \textbf{2.4\%} on NC.

\subsection{Component-wise Ablation Analysis}
Having established the efficacy of the sequential training, we now isolate the impact of granular components within each stage. Specifically, we ablate individual reward terms and the search-constrained thinking strategy to quantify their distinct contributions to the agent's reasoning capabilities.

\begin{figure}[t]
  \centering
  
  % --- 第一排 ---
  \begin{subfigure}[t]{0.48\columnwidth} % 建议稍微改小一点点(0.48)防止挤换行
    \centering
    \includegraphics[width=\linewidth]{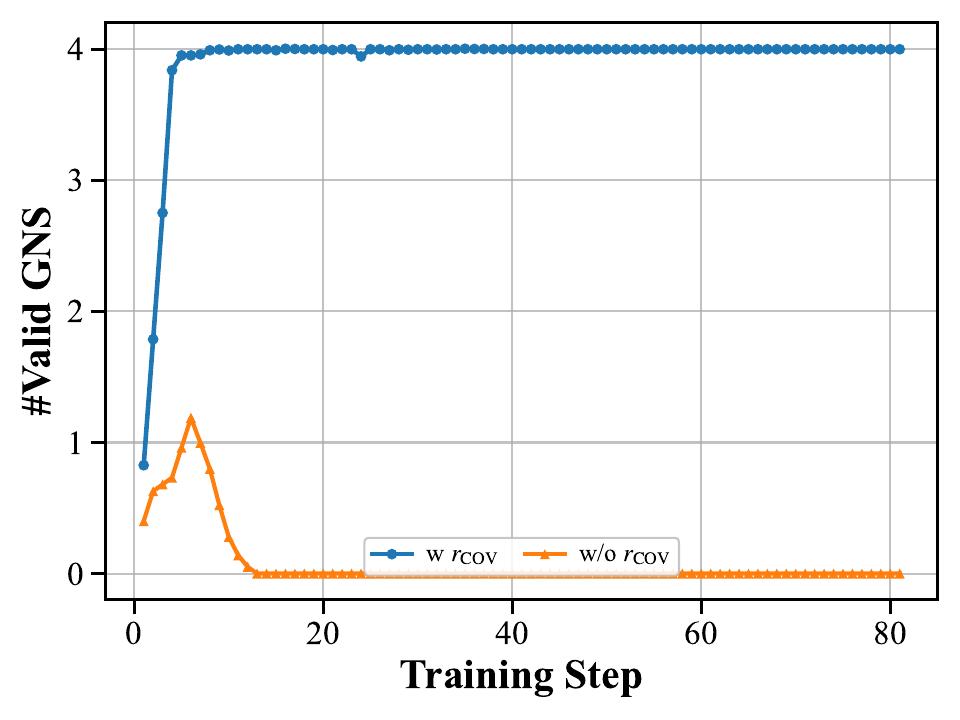}
    \caption{Ablation of $r_{\textsc{cov}}(\tau)$}
    \label{fig:2a}
  \end{subfigure}\hfill % 这一行的左右填充
  \begin{subfigure}[t]{0.48\columnwidth}
    \centering
    \includegraphics[width=\linewidth]{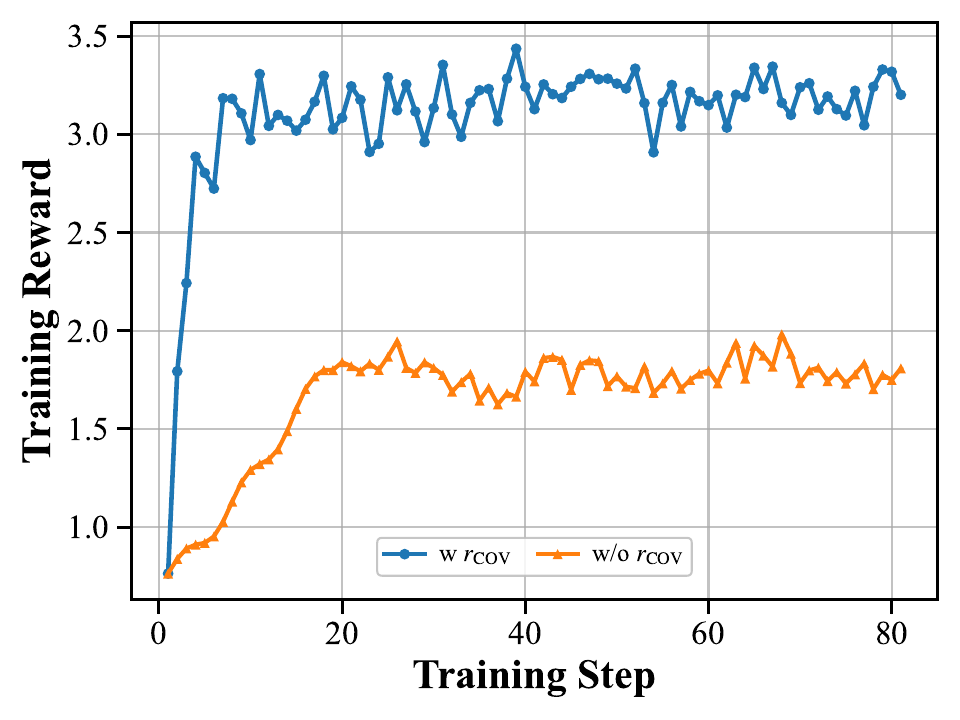}
    \caption{Ablation of $r_{\textsc{cov}}(\tau)$}
    \label{fig:2b}
  \end{subfigure}
  
  \vspace{-1pt} % 可选：增加一点垂直间距，让上下两排不那么挤
  
  % --- 第二排 ---
  \begin{subfigure}[t]{0.48\columnwidth}
    \centering
    \includegraphics[width=\linewidth]{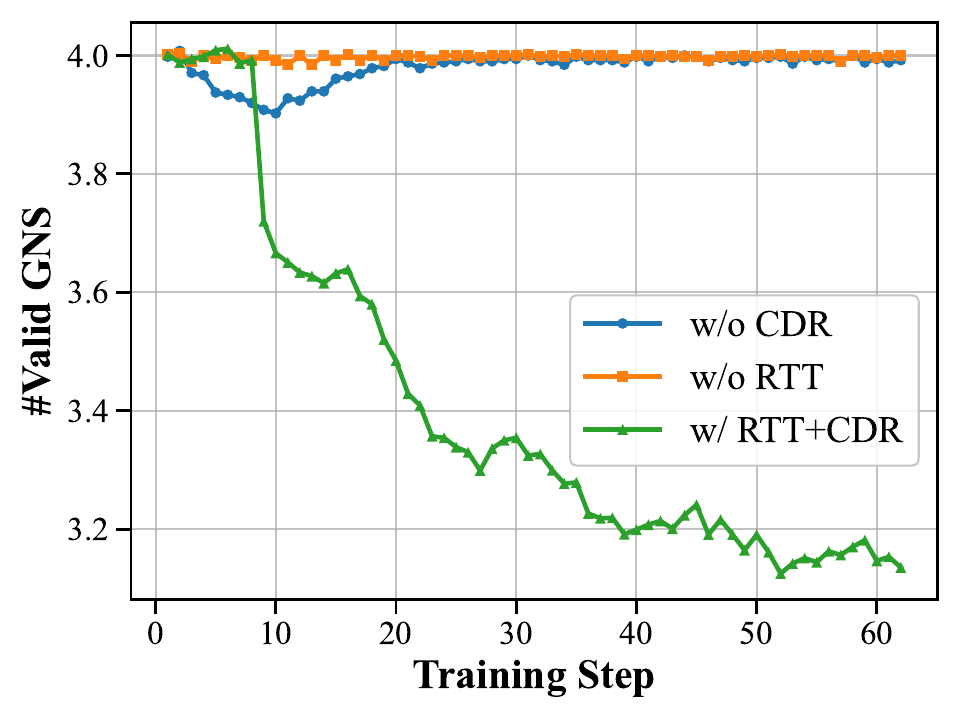} % 替换为你的第三张图
    \caption{Ablation of CDR/RTT}
    \label{fig:2c}
  \end{subfigure}\hfill % 这一行的左右填充
  \begin{subfigure}[t]{0.48\columnwidth}
    \centering
    \includegraphics[width=\linewidth]{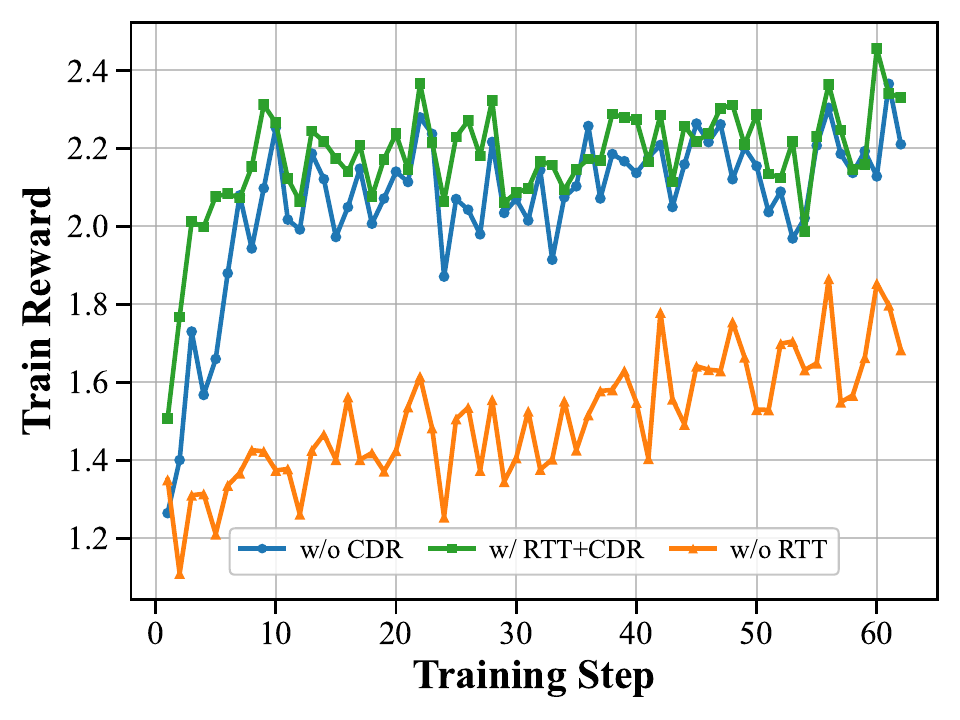} % 替换为你的第四张图
    \caption{Ablation of CDR/RTT}
    \vspace{-1pt}
    \label{fig:method_d}
  \end{subfigure}

  \caption{Ablation study of AgentGL(7B)-GRPO on NC: Analysis of valid GNS counts and training rewards. }
  \label{fig:four_grid}
  \vspace{-2mm}
\end{figure}

% Preamble

\begin{table}[t]
  \centering
  \scriptsize
  \setlength{\tabcolsep}{6pt}
  \renewcommand{\arraystretch}{1.01}

  \begin{tabular}{@{}lccc@{}}
    \specialrule{1.2pt}{0pt}{0pt}
    Model & OGB-Arxiv & Amazon-Photo & Avg. \\
    \specialrule{0.8pt}{2pt}{2pt}

    \textsc{AgentGL} & \textbf{68.9} & \textbf{59.9} & \textbf{64.4} \\
    \quad w/o \textsc{$r_{\textsc{cov}}$} & 65.2 \myred{3.7} & 55.4 \myred{4.5} & 60.3 \myred{4.1} \\
    \quad w/o \textsc{CDR}   & 65.9 \myred{3.0} & 57.1 \myred{2.8} & 61.5 \myred{2.9} \\
    \quad w/o \textsc{RTT}   & 65.4 \myred{3.5} & 56.5 \myred{3.4} & 61.0 \myred{3.4} \\
    \quad w/o \textsc{GCCL}  & 68.2 \myred{0.7} & 59.3 \myred{0.6} & 63.8 \myred{0.6} \\

    \specialrule{1.2pt}{2pt}{0pt}
  \end{tabular}

  \caption{Component-wise ablations of AgentGL. Red numbers denote the absolute improvements.}
  \vspace{-8mm}
  \label{tab:ablation}
\end{table}

\noindent \textbf{\textit{Obs 6. Each component is critical for maintaining the balance between search steps and model performance.}} For Stage 1, as illustrated in Figure~\ref{fig:2a}, in the absence of $r_{\textsc{cov}}(\tau)$, the model fails to acquire effective search habits during training. Consequently, as training steps increase, the agent eventually degenerates to ceasing search operations entirely, maintaining a suboptimal reward level (Figure~\ref{fig:2b}). 
Regarding Stage 2, we conducted an ablation study on the Retrospective Termination Trigger (RTT) and Cognitive Density Regularization (CDR). We observe that without CDR, driven by RTT, the model attempts to improve search efficiency in the early phases of Stage 2; however, this improvement is unsustainable, and the model eventually converges to a search step magnitude similar to that of Stage 1 (Figure~\ref{fig:2c}). Conversely, without RTT, the model persists in the reasoning mode of Stage 1, failing to achieve any efficiency gains. Only the synergistic combination of both components can stably reduce the average search steps, saving approximately \textbf{22\%} of the search cost when training converges while achieving a \textbf{3\%} improvement in accuracy (Table~\ref{tab:ablation}). Furthermore, we perform an ablation study on $\lambda_r$, which governs the embedding-based weighted search for $\tau_{\text{\textsc{1hop}}}$ and $\tau_{\text{\textsc{2hop}}}$. As shown in Table~\ref{tab:lambda_ablation}, the model achieves optimal performance when a balanced weight ($\lambda_r=0.5$) is applied. This observation underscores the necessity of harmonizing structural topology with semantic similarity, which is essential for comprehensive AGL.

\begin{table}[t]
    \centering
    \scriptsize
    % 表格标题：根据需要修改
    
    % 调整列宽，防止太挤
    \setlength{\tabcolsep}{6pt}
    
    % lccc: 第一列左对齐(放lambda)，后三列居中(放数据集数据)
    \begin{tabular}{lccc}
        \toprule
        % 表头：第一列是 Lambda，后三列是数据集名称
        % 请把 Arxiv, PubMed, Amazon 替换为你实际的三个数据集名称
        \multirow{2}{*}{\textbf{Parameter}} & \multicolumn{3}{c}{\textbf{Datasets}} \\
        \cmidrule(lr){2-4}
         & \textbf{OGB-Arxiv} & \textbf{PubMed} & \textbf{Amazon-Computers} \\
        \midrule
        
        % 示例：lambda = 0.0 (Baseline)
        $\lambda_r = 0.0$ & 67.1 & 80.6 & 65.7 \\
        
        % 示例：lambda = 0.5
        $\lambda_r = 0.5$ & \textbf{68.9} & \textbf{82.7} & \textbf{68.6} \\
        
        % 示例：lambda = 1.0
        $\lambda_r = 1.0$ & 66.9 & 80.1 & 66.2 \\
        
        \bottomrule
    \end{tabular}
    \caption{Impact of the hyperparameter $\lambda_r$.}
    \label{tab:lambda_ablation}
    \vspace{-4mm}
\end{table}

\begin{figure}[t]
  \centering
  \begin{subfigure}[t]{0.49\columnwidth}
    \centering
    \includegraphics[width=\linewidth]{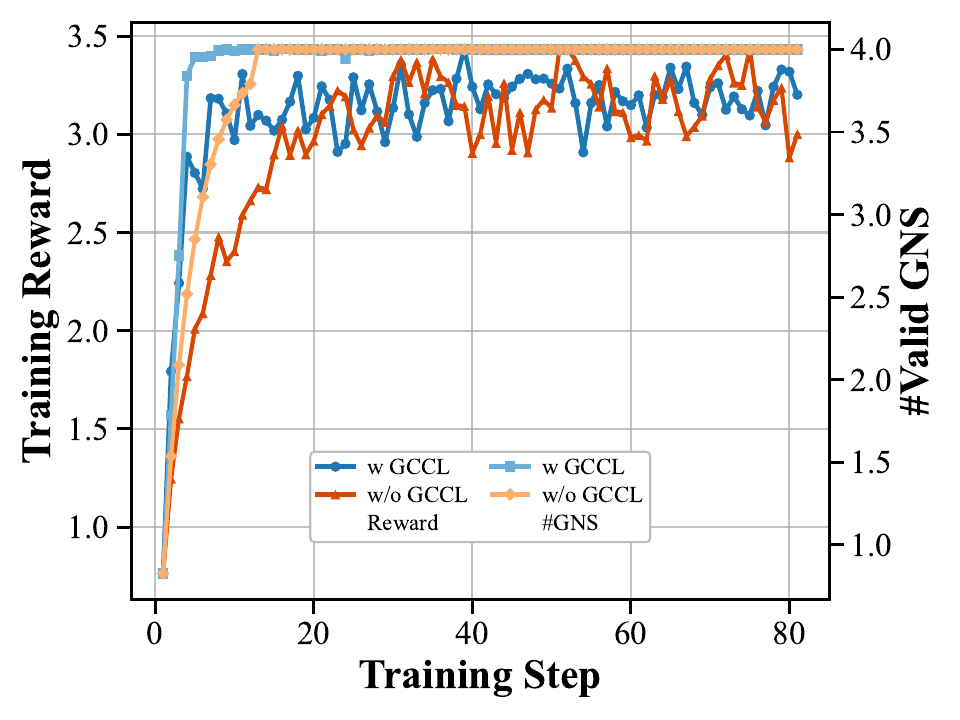}
    \caption{GCCL in Stage1}
    \label{fig:ppo}
  \end{subfigure}\hfill%
  \begin{subfigure}[t]{0.49\columnwidth}
    \centering
    \includegraphics[width=\linewidth]{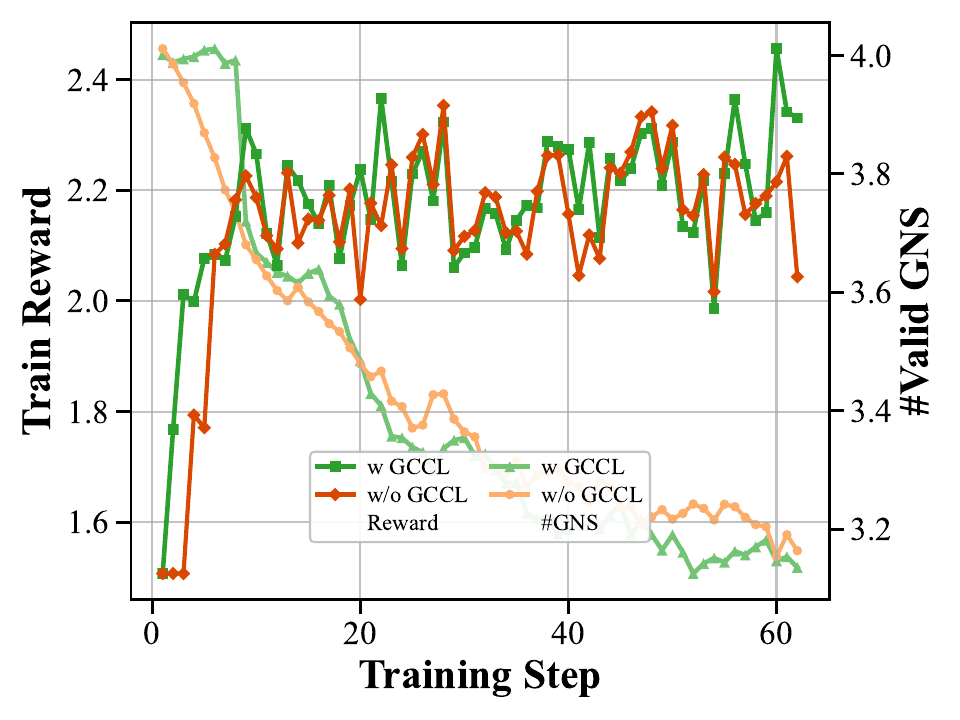}
    \caption{GCCL in Stage2}
    \label{fig:grpo}
  \end{subfigure}
  \vspace{-2mm}

  \caption{Ablation study of GCCL in different stages for AgentGL(7B)-GRPO on NC.}
  \vspace{-6mm}
  \label{fig:3}
\end{figure}

\subsection{Study of Graph-Conditioned Curriculum Learning (GCCL)}
\noindent \textbf{\textit{Obs 7. GCCL serves to stabilize and expedite convergence across distinct training stages.}} 
As illustrated in Figure~\ref{fig:3}, for Stage 1, GCCL effectively accelerates reward convergence and mitigates oscillations in the later phases of training. A similar trend is observed in Stage 2, where GCCL stabilizes the GNS frequency, maintaining the search steps at a consistently lower magnitude compared to the baseline without GCCL as training progresses. Furthermore, quantitative results in Table~\ref{tab:ablation} corroborate that GCCL not only expedites convergence but also yields an accuracy improvement of approximately 0.65\%. In essence, GCCL serves as a stabilizing backbone, effectively guiding the LLM through the complex graph exploration space without succumbing to local optima or early-stage volatility.

%% file: latex/07_conclusion.tex
\section{Conclusion}
% In this paper, we introduce Agentic Graph Learning (AGL), a novel paradigm that reformulates Graph Learning as an interleaved process of graph exploration and reasoning. Building upon this, we propose \texttt{AgentGL}, the first RL-driven framework designed for AGL. 
In this paper, we propose \texttt{AgentGL}, the first RL-driven agentic framework for graph learning, which reformulates Graph Learning as an interleaved process of topology-aware exploration and LLM-based reasoning. \texttt{AgentGL} leverages graph-native search tools for effective navigation and employs a two-stage RL strategy to balance accuracy and efficiency. Across multiple LLM backbones and benchmark settings, \texttt{AgentGL} consistently outperforms strong baselines, including GraphLLMs and GraphRAG methods, achieving the best average performance across all backbones, with absolute gains of up to 17.5\% on node classification and 28.4\% on link prediction. We hope this work inspires further research into agent-based approaches for complex graph reasoning tasks.
% \texttt{AgentGL} is empowered by graph-native search tools for graph navigation, and utilizes a two-stage RL strategy to balance accuracy and efficiency. Experiments demonstrate that \texttt{AgentGL}, across various LLM backbones, significantly outperforms various kinds of baselines, achieving accuracy improvements of 19.6\% in NC and 26.1\% in LP. We hope this work inspires further exploration into agent-based paradigms for tackling complex graph reasoning tasks.

\section{Limitations}
\texttt{AgentGL} currently operates on text-attributed graphs and does not yet support multimodal-attributed graphs, limiting its applicability in settings where nodes contain richer modal information. Moreover, stable performance in the MSO stage depends critically on a careful trade-off in data allocation between the two stages. It also remains worth investigating whether the MSO stage alters the distribution of tool usage during inference time. The MSO stage is designed to be simple, direct, and effective, and we hope it will encourage future research toward more advanced designs. Finally, extending \texttt{AgentGL} to denser graphs also remains a potential direction for future exploration.

\section{Acknowledgments}
We sincerely thank the anonymous reviewers, AC and SAC for their valuable feedback. This research is partially supported by National Natural Science Foundation of China (62402320), Shanghai Science and Technology Program (24YF2731000), and the NYU Shanghai Boost Fund.

%% file: latex/08_appendix.tex
\clearpage
\begin{table*}[t]
\vspace{-4pt}
\centering
\small
\setlength{\tabcolsep}{7pt}
\renewcommand{\arraystretch}{1.15}
\resizebox{\linewidth}{!}{%
\begin{tabular}{l|l|r|r|r}
\hline
\textbf{Domain} & \textbf{Dataset} & \textbf{\#Nodes} & \textbf{\#Edges} & \textbf{\#Classes } \\
\hline\hline
Citation Network & OGB-Arxiv          & 169,343 & 1,166,245 & 40 \\
Citation Network & PubMed            & 19,717  & 44,338    & 3  \\
Citation Network & Arxiv-2023         & 46198 & 78548 & 40 \\
\hline
Amazon Products  & OGB-Products (subset) & 54,025  & 74,420    & 47 \\
Amazon Products  & Amazon-Photo       & 48,362  & 500,939   & 12 \\
Amazon Products  & Amazon-Computers   & 87,229  & 721,107   & 10 \\
\hline
Social Network   & Reddit             & 15,894  & 566,160   & 20 \\
\hline
\end{tabular}}
\vspace{-2pt}
\caption{Dataset statistics used in this paper across three domains. \#Classes correspond to the node classification label space; link prediction is treated as binary.}
\label{tab:dataset_stats}
\vspace{-6pt}
\end{table*}

\section{Appendix}

\subsection{Dataset Details}
\label{app:datasets}

We evaluate AgentGL on 7 text-attributed graph (TAG) benchmarks spanning three domains: \emph{citation networks}, \emph{e-commerce product graphs}, and \emph{social networks}.
Across all TAGs, each node is paired with a piece of natural language text (e.g., title/abstract for papers, product descriptions for items, or post text for online forums), which serves as the semantic grounding for retrieval and reasoning; meanwhile, edges encode native relational structures in the underlying domain, including citation links between papers, co-purchase/co-view relations between products, or interaction/co-posting relations in social platforms.
This combination yields a realistic setting for agentic graph learning: the agent must jointly leverage \emph{topology} (where to search) and \emph{semantics} (what evidence says) to solve downstream tasks, while avoiding redundant evidence accumulation.

On each dataset, we consider two classical graph-learning problems: node classification (predicting a node's category label) and link prediction (predicting whether an edge exists between a pair of nodes).
Node classification evaluates how effectively the model aggregates multi-hop structural cues together with node text to infer labels, whereas link prediction stresses relational reasoning and neighborhood consistency under sparse supervision.
Unless otherwise specified, link prediction is formulated as a binary decision, and node classification uses the original multi-class label space of each dataset.
Table~\ref{tab:dataset_stats} summarizes key statistics, including the number of nodes, edges, and classes.

For data splits, except for Reddit and Arxiv-2023, we follow the default split protocol used in GraphICL \cite{sun2025graphicl} and apply subsampling.
For node classification, on the two training datasets (OGB-Arxiv and OGB-Products), we sample 3,000 training nodes each for optimization, and for each dataset we sample 1,000 nodes from the original test split for evaluation.
For Arxiv-2023, we keep the original split \cite{he2023harnessing} and likewise perform subsampling on its test split for evaluation.
For Reddit, since the original benchmark \cite{yan2025graph} is a multimodal graph, we convert it into a TAG by removing the image attributes of each node and retaining only the original textual fields; we then subsample its test split using the same evaluation protocol. For all datasets, we use Qwen2.5-72B-Instruct to produce concise rewrites of node attributes as a unified preprocessing step. These rewritten attributes are used consistently for AgentGL and all baselines, so as to control context length, and ensure a fair comparison under the same input budget.

\subsection{More Related Work: GraphRAG vs. AGL}
\label{app:graphrag-vs-agl}
GraphRAG-style methods extend classical RAG by incorporating graph structure into evidence selection and organization, typically for open-ended question answering or long-form generation. In this paradigm, a graph (often a knowledge graph) serves as an index or scaffold that helps retrieve and aggregate textual evidence (documents, passages, entity descriptions, or triples) to ground an LLM's generation \cite{jimenez2024hipporag,zhuang2025linearrag}. The core objective is therefore \emph{generation quality} (e.g., factuality, faithfulness, relevance), where the graph is an auxiliary structure that improves retrieval and attribution. In contrast, Agentic Graph Learning (AGL) treats the graph as the \emph{primary problem instance} rather than an external knowledge base. The goal is to solve graph learning/reasoning tasks whose correctness depends on structural signals (e.g., neighborhood composition, multi-hop dependencies, or structural ranking), such as node classification, link prediction, and other graph-native queries. Accordingly, the agent interacts with the environment through \emph{graph-native operators} that return nodes' text attributes, and the episode terminates with a discrete task decision, instead of free-form generation. This framing yields trajectories that are inherently \emph{graph-operational}: the policy learns \emph{which structural context to acquire} under a budget and \emph{when to stop}, rather than retrieving text to write an answer. Although several recent works \cite{luo2025graphr1} have explored agentic GraphRAG, this is not equivalent to agentic graph learning, as the two lines of research focus on fundamentally different objectives. Nevertheless, given their high-level similarities, we select representative (canonical) GraphRAG baselines and adapt them to perform graph reasoning, and we provide a detailed empirical comparison in the experiments section.

\subsection{Implementation Details}
\label{app:implementation}
\paragraph{GraphRAG baselines.}
For HippoRAG2, LinearRAG and GraphCoT, we follow their original settings whenever applicable.
For HippoRAG2, we use gpt-4o-mini for entity extraction and nv-embed-v2 as the embedding model for retrieval.
For LinearRAG, we use spaCy for named entity recognition and all-mpnet-base-v2 for embeddings.
Both configurations match the default choices reported in their respective papers.
To build the retrieval graph in a way that balances downstream task requirements with the original construction scale of these methods, we subsample 500 nodes from each TAG and construct the GraphRAG index using their \emph{original} node text attributes (before any additional processing), which is then used for retrieval during inference.
This subsampled graph is used solely for retrieval/augmentation and is applied consistently across compared methods.
For GraphCoT, we follow the original implementation and prompting protocol. In the experiments, we categorize it as a GraphRAG method because it shares key characteristics with other GraphRAG approaches: it is primarily designed for knowledge-intensive QA, and it supports reasoning by retrieving evidence from a graph augmented with external knowledge.

\paragraph{GraphLLM baselines.}
All GraphLLM baselines are implemented under the same configurations as reported in their original papers, including their default prompting/formatting, graph-to-text serialization strategies, and hyperparameter choices.

\paragraph{GNN baselines.}
For GNN-based baselines, we adopt the same multi-dataset training and transfer protocol as LLaGA \cite{chenllaga}.

\paragraph{Standard Agentic Search Baselines.}
As discussed in the paper, these methods do not natively support search over graph-structured evidence. Moreover, allowing unrestricted online search would make the comparison unfair, since they could directly obtain the answer from the web and bypass any reasoning. Therefore, we keep their original prompting strategies, training protocols, and other settings unchanged, and only replace their online search component with a constrained variant that restricts search to the nodes within the input graph. For Search-R1, we apply GRPO training algorithms.

\paragraph{AgentGL.} 
(Hyper-Parameters) We use OpenRLHF as our primary reinforcement learning training framework.
For node classification, each graph-native search tool returns at most 5 retrieved nodes per call, and we append their node text attributes as evidence to the model context.
Unless otherwise specified, we sample 16 rollouts per prompt and use a single episode per update, with zero warmup.
We set the total training batch size to 128 and the rollout batch size to 32, with a KL regularization coefficient of 0 and a learning rate of 2e-6.
We cap the maximum sequence length at 1600 and sample with temperature 1.0. For all text encoding, we use the RoBERTa-Large encoder (all-roberta-large-v1). In GCCL, we set the standard normal quantile $z$ to 1.96 and $\eta$ to 0.05. All experiments are conducted on a node equipped with 8 NVIDIA H100-80G-SXM5 GPUs and 32 Intel Xeon Platinum 8462Y+ CPU cores (2.8\,GHz). We report the average accuracy over 2 rounds.

\noindent (GCCL Training) For graph-conditioned curriculum learning, we pre-partition the training nodes into three difficulty strata (easy/medium/hard) and allocate a fixed quota to each stage.
For both OGB-Arxiv and OGB-Products, Stage 1 uses 800 easy, 500 medium, and 500 hard samples, while Stage 2 uses 200 easy, 500 medium, and 500 hard samples. Training within each stage is conducted on the allocated data in ascending order of difficulty.

\noindent (Reward Details) In Stage~1, we implement a lightweight reward server that scores each rollout by combining (i) task correctness, (ii) format compliance, (iii) tool-usage coverage. Concretely, we first extract the predicted label and compare it with the gold category using a normalized string match. The classification reward is set to 1.5 for an exact match, 0 for a mismatch, and negative when the answer is missing (-1.0) or the sample index is invalid (-0.5).
We additionally apply a format reward to enforce a clean and machine-parsable trajectory structure. A response receives +0.5 if it contains exactly one \texttt{<think>} block and exactly one \texttt{<answer>} block; otherwise it receives -0.5. We further check that query/document delimiters are well-formed (the numbers of begin/end tags match); this adds +0.1 if consistent and -0.3 otherwise. To prevent leakage of tool I/O into the final prediction, we penalize cases where the answer block contains query/document tags (-0.5), where the answer is overly verbose (more than 12 whitespace-separated tokens, -0.2), or where the answer block contains any residual \texttt{<think>} content (-0.3).
To encourage the agent to explore different graph-native search tools in the bootstrapping stage, we add a search-coverage reward based on which tool tags appear in the rollout. Each distinct tool used contributes +0.5, capped at 2.0 in total.
For Stage~2, we keep the format reward and the classification reward unchanged. The only modification is the cognitive-density reward: if any reasoning segment fails to meet the cognitive density requirement, we apply a penalty of -0.2; otherwise, we add a bonus of +0.5 when all segments satisfy the criterion. The segment-length threshold is set to 100 tokens.
For link prediction, the only change is the number of evidence nodes returned by each tool. Specifically, 1-hop Neighborhood Search and 2-hop Neighborhood Search still return up to 5 nodes per call, while Structure Salience Search returns 2 nodes and Graph Dense Search returns 3 nodes.

\noindent \textbf{AI Usage.} We utilized AI exclusively for proofreading assistance.

\subsection{Additional Experiments}
\paragraph{Variance analysis.}
Table~\ref{tab:variance} reports the performance variance over three independent runs under our RL training setup. Overall, the observed variances are modest across both node classification and link prediction, suggesting that AgentGL training is reasonably stable in practice. A consistent trend is that smaller backbones exhibit larger variance than their larger counterparts (e.g., 3B vs.\ 7B), which aligns with the intuition that smaller models are more sensitive to stochasticity in sampling and policy optimization. This gap is especially noticeable on Amazon-Photo, where the 3B variants show higher variance across both tasks.

\begin{table}[ht]
\vspace{-4pt}
\centering
\scriptsize
\setlength{\tabcolsep}{4pt}
\renewcommand{\arraystretch}{1.15}
\begin{tabular}{l|cc|cc}
\hline
\multirow{2}{*}{Model} & \multicolumn{2}{c|}{OGB-Products} & \multicolumn{2}{c}{Amazon-Photo} \\
& Var(NC) & Var(LP) & Var(NC) & Var(LP) \\
\hline\hline
AgentGL-7B-GRPO & 0.3 & 0.4 & 0.2 & 0.3 \\
AgentGL-3B-GRPO & 0.6 & 0.3 & 0.7 & 1.1 \\
AgentGL-7B-R++ & 0.2 & 0.4 & 0.5 & 0.7 \\
AgentGL-3B-R++ & 0.4 & 0.6 & 0.7 & 0.8 \\
\hline
\end{tabular}
\vspace{-2pt}
\caption{Variance of performance over 3 runs for node classification (NC) and link prediction (LP) on two datasets.}
\label{tab:variance}
\vspace{-6pt}
\end{table}

\paragraph{Sensitivity analysis of K value.}
We further analyze the sensitivity to the neighborhood size $K$ in the main text. As shown in Table~\ref{tab:k_sensitivity}, increasing $K$ from 1 to 5 consistently improves accuracy on both OGB-Arxiv and Amazon-Photo, with the best results achieved at $K{=}5$ (68.9\% and 59.9\%, respectively). This suggests that a moderately expanded neighborhood provides more informative structural context for reasoning. In contrast, further increasing $K$ to 7 leads to a slight performance drop on both datasets, indicating that overly large neighborhoods may introduce redundant or noisy information. Overall, these results show that an appropriate neighborhood size is important for balancing contextual richness and noise, and justify our default choice of $K{=}5$.

\begin{table}[t]
\centering
\small
\caption{Sensitivity analysis of $K$ value (AgentGL-7B-GRPO, Node Classification). }
\label{tab:k_sensitivity}
\begin{tabular}{lcccc}
\toprule
Acc (\%) & $K=1$ & $K=3$ & $K=5$ & $K=7$ \\
\midrule
OGB-Arxiv    & 65.7 & 68.1 & 68.9 & 68.6 \\
Amazon-Photo & 56.2 & 59.1 & 59.9 & 59.7 \\
\bottomrule
\end{tabular}
\end{table}

{ \small
\begin{algorithm*}[t]
\DontPrintSemicolon
\caption{AgentGL: Agentic Graph Learning with RL}
\label{alg:agentgl_training_short}

\SetKwFunction{DiffNC}{CalcDiffNC}
\SetKwFunction{DiffLP}{CalcDiffLP}
\SetKwFunction{Split}{SplitByDifficulty}
\SetKwFunction{Sample}{SampleBatch}
\SetKwFunction{Roll}{Rollout}
\SetKwFunction{Fmt}{FmtReward}
\SetKwFunction{Acc}{AccReward}
\SetKwFunction{Cov}{CoverageReward}
\SetKwFunction{Depth}{DepthReward}
\SetKwFunction{Upd}{UpdatePolicy}

\KwIn{
Graph $\mathcal{G}$; stage data $\mathcal{D}^{(1)},\mathcal{D}^{(2)}$;
tools $\mathcal{S}=\{\tau_{1HOP},\tau_{2HOP},\tau_{SS},\tau_{DENSE}\}$;
ref policy $\pi_{\text{ref}}$; KL coeff $\beta$; budget $B$; rollouts $N$;
GCCL params: $(z,\eta)$ (NC), $\mathrm{Sim}$ (LP);
Stage-1 params: $\lambda_a,\eta_{\text{cov}}$; Stage-2 params: $\lambda_a,\alpha,\lambda_d,\delta$;
RL alg $\mathrm{ALG}\in\{\textsc{GRPO},\textsc{R++}\}$.
}
\KwOut{Optimized $\pi_\theta$.}

\BlankLine
\ForEach(\tcp*[f]{Two RL Stages}){$s\in\{1,2\}$}{
    $\mathcal{D}\gets \mathcal{D}^{(s)}$\;
    \eIf(\tcp*[f]{NC: $S_{NC}(v)=\mathrm{WLB}(\hat p_v,d_v;z)+\eta\log(1+d_v)$}){$x$ is node}{
        $Scores\gets \DiffNC(\mathcal{D},\mathcal{G};z,\eta)$
    }{
        $Scores\gets \DiffLP(\mathcal{D};\mathrm{Sim})$\tcp*{$S_{LP}(e)=y_e\mathrm{Sim}(x_u,x_v)+(1-y_e)(1-\mathrm{Sim}(x_u,x_v))$}
    }
    $\{\mathcal{C}_{\text{E}},\mathcal{C}_{\text{M}},\mathcal{C}_{\text{H}}\}\gets \Split(\mathcal{D},Scores)$\;

    \uIf(\tcp*[f]{\textbf{Stage 1: Graph-Native Search Policy Bootstrapping}}){$s=1$}{
        \tcp{$R(\tau)=r_{\text{FMT}}(\tau)+r_{\text{ACC}}(\hat y,y)+r_{\text{COV}}(\tau)$}
    }\Else{
        \tcp{\textbf{Stage 2: Mitigating Search Overuse} (inject retrospective trigger after each tool call)}
        \tcp{$R(\tau)=r_{\text{FMT}}(\tau)+r_{\text{ACC}}(\hat y,y)+r_{\text{depth}}(z)$ (discard $r_{\text{COV}}$)}
    }

    \ForEach{$k\in\{\text{E},\text{M},\text{H}\}$}{
        \While{stage-$s$ budget not exhausted}{
            $\mathcal{B}\gets \Sample(\mathcal{C}_{k})$\;
            \ForEach{$(x,Q,y)\in\mathcal{B}$}{
                \For{$n\gets 1$ \KwTo $N$}{
                    $\tau^{(n)}\gets \Roll(\pi_\theta,\mathcal{G},\mathcal{S};x,Q,B,s)$\tcp*{$a_t=\langle s_t,q_t\rangle$ or \textsc{ANSWER}}
                    $r_{\text{FMT}}\gets \Fmt(\tau^{(n)})$;\;
                    $r_{\text{ACC}}\gets \Acc(\hat y^{(n)},y)=\lambda_a\mathbf{1}[\hat y^{(n)}=y]$\;
                    \uIf{$s=1$}{
                        $r_{\text{COV}}\gets \Cov(\tau^{(n)})=\eta_{\text{cov}}\sum_{j}\mathbf{1}[\exists t:a_t=\tau_j]$\;
                        $R^{(n)}\gets r_{\text{FMT}}+r_{\text{ACC}}+r_{\text{COV}}$\;
                    }\Else{
                        $r_{\text{depth}}\gets \Depth(\tau^{(n)};\alpha,\lambda_d,\delta)=\alpha\mathbf{1}[N_{\text{short}}=0]-\lambda_dN_{\text{short}}$\;
                        $R^{(n)}\gets r_{\text{FMT}}+r_{\text{ACC}}+r_{\text{depth}}$\;
                    }
                }
            }
            $\theta\gets \Upd(\theta,\{\tau^{(n)},R^{(n)}\},\pi_{\text{ref}},\beta;\mathrm{ALG})$\tcp*{GRPO/R++ w/ KL}
        }
    }
}
\Return{$\theta$}\;
\end{algorithm*}
}

\subsection{Case Study}

To make AgentGL's decision process more interpretable, we present representative rollouts for both node classification and link prediction.
Fig.~\ref{fig:nc_case_tripod} shows an NC example from the Amazon domain: the model first forms a hypothesis from the anchor text, then verifies it by querying local neighborhoods (1-hop/2-hop) and a global prior (PageRank).
We highlight the key reasoning sentences that directly support the final label, illustrating how evidence aggregation over the graph reduces ambiguity and prevents over-reliance on the anchor text alone.
Fig.~\ref{fig:lp_case_reddit_bird} provides an LP example from Reddit, where the model validates a potential edge by searching common 1-hop neighbors; the shared co-post motif offers strong structural evidence that the two endpoints lie in the same tight cluster.
Across cases, the agent typically terminates early once the searched evidence becomes self-consistent, avoiding redundant searches under a bounded budget.

% In your preamble, make sure you have (or equivalent):
% \usepackage{listings}
% \usepackage{xcolor}
% \usepackage{ulem} % for \uline (optional)
%
% And you already have the casebox environment defined somewhere, e.g.:
% \newtcolorbox{casebox}[1]{...}

% ---------- Casebox: NC Case Study ----------
\begin{figure*}[t]
\centering
\caption{A node classification (NC) case illustrating graph-native tool use. We highlight key reasoning sentences that drive the final decision.}
\label{fig:nc_case_tripod}
\begin{casebox}{Case Study: Node Classification (NC)}

\textbf{Target instance text attribute:}\\[1pt]
Tripod is not sturdy enough for heavy use; legs bulge and move, hinge is weak. Not recommended for rifles or kneeling shots.

\vspace{2pt}

\textbf{Ground truth label:}\\[1pt]
\textbf{Tripods \& Monopods}

\vspace{2pt}

\textbf{Model prediction:}\\[1pt]
\textbf{Tripods \& Monopods} \quad \textbf{(\cmark\ Matched)}

\vspace{2pt}

\textbf{Model response:}\\[1pt]
\begingroup
\setlength{\fboxsep}{0pt}
\setlength{\fboxrule}{0pt}
\begin{minipage}{0.985\linewidth}
\fontsize{6.7}{7.6}\selectfont

\texttt{<think>}\\
\hl{The anchor summary is already strongly indicative of a tripod: it talks about leg sections bulging, the legs shifting under load, and a weak hinge, plus mentions kneeling shots / rifles (a common tripod/spotting-scope/shooting-use context).}
I’ll still start with local graph evidence and only expand if anything looks like it could actually be an accessory rather than the tripod itself.

\vspace{2pt}
Round 1 — 1-hop:
\hl{Make the query reflect both the mechanical failure points (hinge/legs) and the use case (kneeling/shooting) so the nearest co-purchases are maximally revealing.}

\vspace{2pt}
\texttt{<|begin\_of\_query|>1-hop:tripod weak hinge legs bulge unstable kneeling shooting<|end\_of\_query|>}\\
\texttt{<|begin\_of\_documents|>}\\
\texttt{(1) Budget tripod; quick leg locks; okay for light cameras but wobbles when fully extended.}\\
\texttt{(2) Tripod for spotting scope / range; stable at low height, awkward for kneeling positions.}\\
\texttt{(3) Lightweight travel tripod with ball head; users mention leg flex and hinge durability issues.}\\
\texttt{(4) “Heavy duty” tripod claims; complaints about loose joints and bulging leg sections under load.}\\
\texttt{(5) Compact aluminum tripod; fine for occasional use, not recommended for heavy rifles or heavy camera rigs.}\\
\texttt{<|end\_of\_documents|>}\\
\vspace{2pt}
\hl{These direct neighbors read like the same product family: tripods criticized for stability, load rating, and joint/hinge reliability—plus the same shooting/kneeling angle shows up.}
That’s enough to lean hard toward “Tripods \& Monopods,” but I’ll expand once to check whether the node could be a tripod component (e.g., a head/plate) that’s being discussed with tripod-like language.

\vspace{2pt}
Round 2 — 2-hop:

\hl{Probe for whether the neighborhood shifts toward “plates/heads/clamps” as the main object, or stays centered on the support legs/column system.}

\vspace{2pt}
\texttt{<|begin\_of\_query|>2-hop:tripod load capacity leg locks center column wobble head loosens<|end\_of\_query|>}\\
\texttt{<|begin\_of\_documents|>}\\
\texttt{(1) Tripod head loosens and tilt lock slips; fine for phones, not for heavier cameras.}\\
\texttt{(2) Tripod legs slip under load; center column wobble; advertised rating overstated.}\\
\texttt{(3) Monopod suggested as alternative; compared directly against tripods for portability.}\\
\texttt{(4) Video tripod: smooth pan but poor tilt stability for heavier rigs.}\\
\texttt{(5) Travel tripod: stable low, unstable fully extended; needs added weight for stability.}\\
\texttt{<|end\_of\_documents|>}\\

\vspace{2pt}
\hl{Still tripod-first: legs, center column, extension stability, and load capacity dominate the neighborhood, with monopod comparisons appearing naturally.}
Nothing suggests the anchor is primarily a quick-release plate or a clamp.
One global pass can help ensure we’re not missing a category boundary in the label set, so I’ll do a pagerank lookup.

\vspace{2pt}
Round 3 — pagerank:

\hl{Use terms that usually surface canonical support gear (tripods/monopods/heads/plates) and see what comes back.}

\vspace{2pt}
\texttt{<|begin\_of\_query|>pagerank:tripod monopod tripod head quick release plate travel tripod<|end\_of\_query|>}\\
\texttt{<|begin\_of\_documents|>}\\
\texttt{(1) Popular entry-level tripod: quick-release plate, bubble level, weight hook.}\\
\texttt{(2) Common ball head / tripod head frequently paired with tripods.}\\
\texttt{(3) Tripod carry bag / case often co-purchased for transport.}\\
\texttt{(4) Universal quick-release plate accessory used across tripod brands.}\\
\texttt{(5) Well-known travel tripod line referenced in comparisons (compact fold, lightweight).}\\
\texttt{<|end\_of\_documents|>}\\

\vspace{2pt}
\hl{This is exactly the ecosystem we’d expect when the anchor is a tripod: the graph surfaces other tripods and the usual companion pieces (head/plate/bag), but the core story remains “support legs + stability under load,” matching the anchor summary.}
{\sethlcolor{green!18}\hl{A fourth “similar” search would likely be redundant, so I’ll stop here.}}

\texttt{</think>}\\[1pt]
\texttt{<answer>Tripods \& Monopods</answer>}
\end{minipage}
\endgroup

\end{casebox}
\vspace{-10pt}
\end{figure*}

\begin{figure*}[t]
\centering
\caption{A link prediction (LP) case from Reddit. The model verifies a strong co-post motif by retrieving dense common 1-hop neighbors.}
\label{fig:lp_case_reddit_bird}
\begin{casebox}{Case Study: Link Prediction (LP) --- Reddit}

\textbf{Target instance text attribute:}\\[2pt]
\textit{Node U:} laggar falcon - falco jugger - tal chappar, rajasthan, india\\[2pt]
\textit{Node V:} greater hoopoe-lark : the courtship display of the male consists of rising with fluttering wing-strokes and then diving down with closed wings to a perch. - banni grassland, gujarat, india

\medskip

\textbf{Ground truth label:}\\[2pt]
\textbf{yes}

\medskip

\textbf{Model prediction:}\\[2pt]
\textbf{yes} \quad \textbf{(\cmark\ Matched)}

\medskip\medskip

\textbf{Model response:}\\[2pt]
\begingroup
\begin{minipage}{0.985\linewidth}
\scriptsize

\texttt{<think>}\\
\hl{U and V are both bird-observation style Reddit posts with the same “species -- location -- India” pattern.}
\hl{The fastest way to verify that is to check for common 1-hop neighbors: if they share multiple direct neighbors that are themselves highly consistent, that’s strong evidence of an edge.}

\medskip
Round 1 --- 1-hop (prioritize common direct neighbours):\\[2pt]

\texttt{<|begin\_of\_query|>1-hop:birding india tal chappar rajasthan banni grassland gujarat laggar falcon hoopoe-lark<|end\_of\_query|>}\\

\texttt{<|begin\_of\_documents|>}\\
\texttt{(1) [common one\_hop neighbour] western marsh harrier - little rann of kutch, gujarat, india This neighbour links both Node U and Node V.}\\
\texttt{(2) [common one\_hop neighbour] peregrine falcon - little rann of kutch, gujarat, india This neighbour links both Node U and Node V.}\\
\texttt{(3) [common one\_hop neighbour] common babbler - little rann of kutch, gujarat, india This neighbour links both Node U and Node V.}\\
\texttt{(4) [common one\_hop neighbour] baya weaver - lodai, kutch, gujarat, india This neighbour links both Node U and Node V.}\\
\texttt{(5) [common one\_hop neighbour] short-eared owl - asio flammeus - little rann of kutch, gujarat, india This neighbour links both Node U and Node V.}\\
\texttt{<|end\_of\_documents|>}\\

\medskip
\hl{They share a dense set of direct common neighbors, all of which are the same type of birding posts from the same west-India region (Kutch / Gujarat) and the same naming template.}
{\sethlcolor{green!18}\hl{This is a strong co-post motif in the Reddit graph, indicating U and V belong to the same closely connected cluster.}}

\texttt{</think>}\\[2pt]
\texttt{<answer>yes</answer>}
\end{minipage}
\endgroup

\end{casebox}
\vspace{-6pt}
\end{figure*}

\subsection{Prompt Template}
We use a standardized prompt format to expose graph-native search tools to the policy in a machine-parsable manner.
Figs.~\ref{fig:nc_prompt_core}--\ref{fig:lp_prompt_limits} summarize the templates used for node classification and link prediction.
Each prompt consists of a \texttt{system} instruction specifying the task, the available search pools, and strict output constraints, followed by a \texttt{user} message that provides the anchor instance (NC) or a node pair (LP).
Placeholders \{\{...\}\} are instantiated with dataset-specific task lines, label spaces, relation descriptions, and per-pool Top-$K$ limits, while keeping the action interface invariant across domains.
During inference (and RL rollouts), all tool calls must appear inside a single \texttt{<think>} block, and the final prediction must be emitted as exactly one label enclosed by \texttt{<answer>} tags.
This design ensures consistent trajectory logging, robust automatic reward computation, and reproducible evaluation under a fixed search budget.
% ---------------------------
% 1) Main NC template (core)
% ---------------------------
\begin{figure*}[t]
\centering
\caption{NC prompt template (core). Placeholders in the prompt use \{\{...\}\}.}
\label{fig:nc_prompt_core}
\begin{caseboxpurple}{Prompt Template: Node Classification (Core)}
\scriptsize
\begin{lstlisting}[basicstyle=\ttfamily\scriptsize, breaklines=true]

<|im_start|>system
You are a reasoning assistant with the ability to perform graph searches to help you answer the user's question accurately.
{{TASK_LINE}}
You can use graph search to retrieve neighbor node information to support your prediction.

THINK/ANSWER FORMAT:
- Do ALL internal reasoning inside <think>...</think>.
- Provide ONLY the final category label inside <answer>...</answer>.
- Never reveal your chain-of-thought outside <think>...</think>.
- When confident, output exactly one label as required.

GRAPH SEARCH POOLS (for context retrieval):
{{POOLS}}

SEARCH FORMAT ({{SEARCH_LIMITS_DESC}}):
- All searches MUST be performed INSIDE the <think>...</think> block.
- Use one of the pool-specific tags below:
  * 1-hop:    <|begin_of_query|>1-hop:QUERY<|end_of_query|>
  * 2-hop:    <|begin_of_query|>2-hop:QUERY<|end_of_query|>
  * pagerank: <|begin_of_query|>pagerank:QUERY<|end_of_query|>
  * similar:  <|begin_of_query|>similar:QUERY<|end_of_query|>
- Results are returned as:
  <|begin_of_documents|> ... <|end_of_documents|>

GUIDELINES:
- One search per round; multiple rounds allowed.
- Max total searches = {{MAX_SEARCH_LIMIT}}.
- Output only <think>...</think> and a single <answer>...</answer>.

OUTPUT REQUIREMENT:
{{OUTPUT_REQ}}
<|im_end|>

<|im_start|>user
{{ANCHOR_HDR}}Anchor Node Information:
{{SUMMARY_SNIPPET}}
<|im_end|>

<|im_start|>assistant
\end{lstlisting}
\end{caseboxpurple}
\vspace{-6pt}
\end{figure*}

% --------------------------------
% 2) Dataset-specific inserts: arxiv
% --------------------------------
\begin{figure*}[t]
\centering
\caption{Dataset-specific inserts for Arxiv (NC).}
\label{fig:nc_prompt_arxiv}
\begin{caseboxpurple}{Dataset Inserts: Arxiv}
\scriptsize
\begin{lstlisting}[basicstyle=\ttfamily\scriptsize, breaklines=true]

TASK_LINE:
Your task is to predict the category of arXiv computer science (cs) papers.

OUTPUT_REQ:
- Final output must be exactly: <answer>cs.XX</answer>

POOLS:
- 1-hop: direct neighbors (papers that directly cite or are cited by the anchor). Returns up to {{TOPK_ONE_HOP}} nodes.
- 2-hop: neighbors of neighbors expanding the local region. Returns up to {{TOPK_TWO_HOP}} nodes.
- pagerank: globally influential nodes selected by PageRank. Returns up to {{TOPK_PAGERANK}} nodes.
- similar: globally most semantically similar nodes by embedding similarity. Returns up to {{TOPK_SIMILAR}} nodes.

ANCHOR_HDR:
Now please predict the category of the anchor node paper:

SEARCH_LIMITS_DESC:
per-pool limits -> 1-hop {{TOPK_ONE_HOP}}, 2-hop {{TOPK_TWO_HOP}}, pagerank {{TOPK_PAGERANK}}, similar {{TOPK_SIMILAR}}
\end{lstlisting}
\end{caseboxpurple}
\vspace{-6pt}
\end{figure*}

% ---------------------------------
% 3) Dataset-specific inserts: pubmed
% ---------------------------------
\begin{figure*}[t]
\centering
\caption{Dataset-specific inserts for PubMed (NC).}
\label{fig:nc_prompt_pubmed}
\begin{caseboxpurple}{Dataset Inserts: PubMed}
\scriptsize
\begin{lstlisting}[basicstyle=\ttfamily\scriptsize, breaklines=true]

TASK_LINE:
Your task is to predict the category of PubMed biomedical papers.

OUTPUT_REQ:
- Final output must be exactly one of the listed PubMed categories inside <answer>...</answer>.

POOLS:
- 1-hop: direct neighbors (papers that directly cite or are cited by the anchor). Returns up to {{TOPK_ONE_HOP}} nodes.
- 2-hop: neighbors of neighbors expanding the local region. Returns up to {{TOPK_TWO_HOP}} nodes.
- pagerank: globally influential nodes selected by PageRank. Returns up to {{TOPK_PAGERANK}} nodes.
- similar: globally most semantically similar nodes by embedding similarity. Returns up to {{TOPK_SIMILAR}} nodes.

ANCHOR_HDR:
Now please predict the category of the anchor node paper:
Available PubMed categories:
- {{CATEGORY_LIST}}

SEARCH_LIMITS_DESC:
per-pool limits -> 1-hop {{TOPK_ONE_HOP}}, 2-hop {{TOPK_TWO_HOP}}, pagerank {{TOPK_PAGERANK}}, similar {{TOPK_SIMILAR}}
\end{lstlisting}
\end{caseboxpurple}
\vspace{-6pt}
\end{figure*}

% --------------------------------
% 4) Dataset-specific inserts: amazon
% --------------------------------
\begin{figure*}[t]
\centering
\caption{Dataset-specific inserts for Amazon (NC).}
\label{fig:nc_prompt_amazon}
\begin{caseboxpurple}{Dataset Inserts: Amazon}
\scriptsize
\begin{lstlisting}[basicstyle=\ttfamily\scriptsize, breaklines=true]

TASK_LINE:
Your task is to predict the category of Amazon products.

OUTPUT_REQ:
- Final output must be exactly one of the listed Amazon categories inside <answer>...</answer>.

POOLS:
- 1-hop: direct neighbors (products that are frequently co-purchased with the anchor). Returns up to {{TOPK_ONE_HOP}} nodes.
- 2-hop: neighbors of neighbors expanding the local region in the co-purchasing graph. Returns up to {{TOPK_TWO_HOP}} nodes.
- pagerank: globally influential products selected by PageRank. Returns up to {{TOPK_PAGERANK}} nodes.
- similar: globally most semantically similar products by embedding similarity. Returns up to {{TOPK_SIMILAR}} nodes.

ANCHOR_HDR:
Now please predict the category of the anchor node product:
Available Amazon categories:
- {{CATEGORY_LIST}}

SEARCH_LIMITS_DESC:
per-pool limits -> 1-hop {{TOPK_ONE_HOP}}, 2-hop {{TOPK_TWO_HOP}}, pagerank {{TOPK_PAGERANK}}, similar {{TOPK_SIMILAR}}
\end{lstlisting}
\end{caseboxpurple}
\vspace{-6pt}
\end{figure*}

% --------------------------------
% 5) Dataset-specific inserts: reddit
% --------------------------------
\begin{figure*}[t]
\centering
\caption{Dataset-specific inserts for Reddit (NC).}
\label{fig:nc_prompt_reddit}
\begin{caseboxpurple}{Dataset Inserts: Reddit}
\scriptsize
\begin{lstlisting}[basicstyle=\ttfamily\scriptsize, breaklines=true]

TASK_LINE:
Your task is to predict the subreddit category of a Reddit post.

OUTPUT_REQ:
- Final output must be exactly one of the listed subreddit categories inside <answer>...</answer>.

POOLS:
- 1-hop: other posts from the same author (also_posted set). Returns up to {{TOPK_ONE_HOP}} nodes.
- 2-hop: same as 1-hop in this dataset (duplicates allowed). Returns up to {{TOPK_TWO_HOP}} nodes.
- pagerank: globally influential posts selected by PageRank over the author cliques. Returns up to {{TOPK_PAGERANK}} nodes.
- similar: globally most semantically similar posts by embedding similarity. Returns up to {{TOPK_SIMILAR}} nodes.

ANCHOR_HDR:
Now please predict the category of the anchor post:
Available subreddit categories:
- {{CATEGORY_LIST}}

SEARCH_LIMITS_DESC:
per-pool limits -> 1-hop {{TOPK_ONE_HOP}}, 2-hop {{TOPK_TWO_HOP}}, pagerank {{TOPK_PAGERANK}}, similar {{TOPK_SIMILAR}}
\end{lstlisting}
\end{caseboxpurple}
\vspace{-6pt}
\end{figure*}

% Preamble requirements:
% \usepackage{listings}
% (Recommended) \lstset{columns=fullflexible, keepspaces=true}

% ---------------------------
% 1) Main LP template (core)
% ---------------------------
\begin{figure*}[t]
\centering
\caption{Link prediction (LP) prompt template (core). Placeholders in the prompt use \{\{...\}\}.}
\label{fig:lp_prompt_core}
\begin{caseboxpurple}{Prompt Template: Link Prediction (Core)}
\scriptsize
\begin{lstlisting}[basicstyle=\ttfamily\scriptsize, breaklines=true]

<|im_start|>system
You are a reasoning assistant with access to graph searches.
Determine whether two nodes from the {{DATASET}} dataset should be connected.
{{RELATION_DESC}}
Treat the task as binary classification and return 'yes' if the edge should exist and 'no' otherwise.

THINK/ANSWER FORMAT:
- Perform all reasoning and search planning inside <think>...</think>.
- Output ONLY the final judgment as <answer>yes</answer> or <answer>no</answer>.
- Never leak your chain-of-thought outside of <think>...</think>.

GRAPH SEARCH POOLS:
- 1-hop: prioritize common direct neighbours (connected to BOTH endpoints U and V). If insufficient, fill with non-common 1-hop neighbours from U and/or V (balanced when possible).
- 2-hop: prioritize common 2-hop neighbours (reachable from BOTH endpoints within two hops). If insufficient, fill with non-common 2-hop neighbours from U and/or V (balanced when possible).
- pagerank: list globally influential reference edges selected offline using PageRank, as complementary structural priors.
- similar: retrieve Top-K node pairs most similar to the current pair, including their edge status as reference.

SEARCH FORMAT ({{SEARCH_LIMITS_DESC}}):
- Every search must happen inside <think>...</think>.
- To launch a search, emit exactly one of the following tags with an optional free-form hint:
  * <|begin_of_query|>1-hop:Query<|end_of_query|>
  * <|begin_of_query|>2-hop:Query<|end_of_query|>
  * <|begin_of_query|>pagerank:Query<|end_of_query|>
  * <|begin_of_query|>similar:Query<|end_of_query|>
- Retrieved blocks arrive as <|begin_of_documents|> ... <|end_of_documents|>.
- Use at most one search per round and no more than {{MAX_SEARCH_LIMIT}} total searches before answering.

GUIDELINES:
- Start by analyzing the two node descriptions.
- Prefer covering multiple pools (shared neighbours, unique structure, global priors).
- Keep the final judgement concise.
- Final answer MUST be <answer>yes</answer> or <answer>no</answer>.
<|im_end|>

<|im_start|>user
We are investigating whether the following two nodes should be linked:
- Node U (id={{NODE_U}}): {{SUMMARY_U}}
- Node V (id={{NODE_V}}): {{SUMMARY_V}}

Use up to {{MAX_SEARCH_LIMIT}} searches to gather evidence from the graph retriever.
- Label 1 corresponds to <answer>yes</answer>.
- Label 0 corresponds to <answer>no</answer>.
Remember to answer strictly with yes/no enclosed by <answer> tags.
<|im_end|>

<|im_start|>assistant
\end{lstlisting}
\end{caseboxpurple}
\vspace{-6pt}
\end{figure*}

% -----------------------------------------
% 2) Dataset relation descriptions (separate)
% -----------------------------------------
\begin{figure*}[t]
\centering
\caption{Dataset-specific relation descriptions used in LP prompts.}
\label{fig:lp_prompt_relations}
\begin{caseboxpurple}{Dataset Inserts: Relation Descriptions (LP)}
\scriptsize
\begin{lstlisting}[basicstyle=\ttfamily\scriptsize, breaklines=true]

[arxiv/pubmed]
Nodes are research papers. An edge represents a citation linkage between the two papers.

[amazon/products]
Nodes are Amazon products. An edge indicates strong co-purchase relationships between the items.

[reddit]
Nodes are Reddit posts. An edge indicates strong co-post relationships between the posts.

[default]
Nodes come from the same graph dataset; an edge captures the canonical relation defined for that dataset.
\end{lstlisting}
\end{caseboxpurple}
\vspace{-6pt}
\end{figure*}

% ----------------------------------------
% 3) Search limits description (separate)
% ----------------------------------------
\begin{figure*}[t]
\centering
\caption{Per-pool search limits description used in LP prompts.}
\label{fig:lp_prompt_limits}
\begin{caseboxpurple}{Dataset Inserts: Search Limits (LP)}
\scriptsize
\begin{lstlisting}[basicstyle=\ttfamily\scriptsize, breaklines=true]

SEARCH_LIMITS_DESC:
per-pool limits -> 1-hop {{TOPK_ONE_HOP}}, 2-hop {{TOPK_TWO_HOP}}, pagerank {{TOPK_PAGERANK}}, similar {{TOPK_SIMILAR}}
\end{lstlisting}
\end{caseboxpurple}
\vspace{-6pt}
\end{figure*}